\documentclass[sigconf]{acmart}
\usepackage{booktabs}        
\usepackage{multirow}        
\usepackage{array}           
\usepackage{makecell}        
\usepackage{ulem}            
\usepackage{bm}              
\usepackage[table]{xcolor}   
\usepackage{xcolor}
\usepackage{ulem}
\usepackage{enumitem}
\usepackage{subcaption}
\usepackage{graphicx}
\usepackage{appendix}
\usepackage{caption}
\usepackage{pifont} 
\usepackage{marvosym}
\usepackage{arydshln}
\usepackage{tabularx}
\usepackage{makecell}
\usepackage{amsmath}
\usepackage{algorithm}
\usepackage{algpseudocode}
\usepackage{multicol}
\usepackage{placeins}
\usepackage{capt-of}
\usepackage[most]{tcolorbox}
\usepackage{listings}
\usepackage{xcolor}

\definecolor{MagicPromptBlue}{RGB}{0,45,115}

\lstdefinestyle{magicpromptstyle}{
    basicstyle=\ttfamily\footnotesize,
    columns=fullflexible,
    keepspaces=true,
    showstringspaces=false,
    breaklines=true,
    breakatwhitespace=true,
    tabsize=2,
    moredelim=[s][\color{MagicPromptBlue}]{\{}{\}},
    moredelim=[is][\bfseries]{**}{**},
    literate={—}{{\textemdash}}1
}

\newtcblisting{magicprompt}[1]{
    enhanced,
    breakable,
    listing only,
    listing engine=listings,
    listing options={style=magicpromptstyle},
    colback=gray!3,
    colframe=gray!55,
    boxrule=0.5pt,
    arc=1mm,
    left=5pt,
    right=5pt,
    top=8pt,
    bottom=5pt,
    before skip=7pt,
    after skip=9pt,
    coltitle=white,
    fonttitle=\footnotesize\bfseries,
    attach boxed title to top left={
        xshift=7pt,
        yshift=-2pt
    },
    boxed title style={
        colback=gray!65,
        colframe=gray!65,
        boxrule=0pt,
        sharp corners
    },
    title={#1}
}

\definecolor{rewardcolor}{RGB}{230,145,0}
\definecolor{advcolor}{RGB}{0,145,95}
\definecolor{creditcolor}{RGB}{0,100,190}
\definecolor{policycolor}{RGB}{175,0,115}
\usepackage{pifont}
\usepackage[table]{xcolor}

\definecolor{TableGray}{HTML}{F2F2F2}
\definecolor{TablePink}{HTML}{FCEAE7}
\definecolor{CheckRed}{HTML}{E3423F}
\definecolor{CrossTeal}{HTML}{34968E}
\definecolor{RefBlue}{HTML}{EAF2F5}

\usepackage{CJKutf8}

\usepackage[dvipsnames]{xcolor}
\definecolor{lightgreen}{RGB}{204, 255, 204}  
\definecolor{lightred}{RGB}{255,204,204}
\definecolor{skyblue}{RGB}{135,206,250}

\AtBeginDocument{%
    \setlength{\abovedisplayskip}{2.5pt}
    \setlength{\belowdisplayskip}{2.5pt}
    \setlength{\abovedisplayshortskip}{2.5pt}
    \setlength{\belowdisplayshortskip}{2.5pt}
}
\AtBeginDocument{%
  }

\copyrightyear{2026}
\acmYear{2026}
\setcopyright{cc}
\setcctype{by}

\begin{document}
\title{Learning to Reason with Predictive Gains for Generative Recommendation}
\author{%
  Zhuoyang Li\textsuperscript{1},
  Teng Shi\textsuperscript{1},
  Jiakai Tang\textsuperscript{1},
  Weijie Yu\textsuperscript{2},
  Guofu Xie\textsuperscript{1},
  Xiao Zhang\textsuperscript{1},
  Lei Kou\textsuperscript{3},\\
  Yang Yu\textsuperscript{3},
  Lei Shen\textsuperscript{3},
  Xin Chen\textsuperscript{3},
  Bing Wang\textsuperscript{3},
  Xiaoyi Zeng\textsuperscript{3},
  Jun Xu\textsuperscript{1 \Letter}
}
\authornote{\Letter Corresponding author.}

\affiliation{%
  \institution{%
    \textsuperscript{1}Gaoling School of Artificial Intelligence,
    Renmin University of China, Beijing \country{China}\\
    \textsuperscript{2}School of Artificial Intelligence and Data Science,
    University of International Business and Economics, Beijing, China\\
    \textsuperscript{3}Alibaba International Digital Commercial Group,
    Hangzhou, China
  }
}
\email{\{zhuoyangli\_03, junxu\}@ruc.edu.cn}

\renewcommand{\shortauthors}{Zhuoyang Li et al.}

\renewcommand{\shortauthors}{Li et al.}

\begin{abstract}
Reasoning-enhanced generative recommendation (GR) has attracted increasing attention for leveraging the reasoning capabilities of large language models (LLMs) to improve next-item prediction. However, existing methods primarily optimize reasoning through Group Relative Policy Optimization (GRPO) with rule-based rewards determined by final recommendation performance, which faces two key limitations: (i) unmeasured predictive contributions within reasoning, where the predictive contributions of individual reasoning steps are not directly reflected in optimization signals, and (ii) weakly discriminative learning signals within  groups, where different reasoning trajectories within the group often receive indistinguishable rewards, limiting effective policy optimization. In this paper, we propose \textbf{GainReason}, a simple yet effective training framework that optimizes reasoning through  \textbf{Predictive Gain (PG)}. Specifically, PG measures changes in target Semantic ID (SID) log-likelihood induced by individual reasoning sentences, providing contribution-aware and discriminative rewards. We further theoretically establish the alignment between expected PG and the recommendation ranking objective. Building on this signal, we develop PG-based Reasoning Optimization (PGRO) to optimize reasoning generation with sentence-level rewards, supported by a custom causal attention mechanism for efficient PG computation. Additionally, an auxiliary PG-based Reasoning Utilization (PGRU) task adaptively strengthens reasoning-conditioned prediction to complement PGRO. Extensive experiments on three public benchmarks demonstrate that GainReason consistently outperforms competitive baselines and achieves substantial performance gains from reasoning. Our codes are publicly available. \footnote{ \url{https://github.com/easonzyli/GainReason}}
\end{abstract}




\maketitle
 \section{Introduction}
\begin{figure}[t]
    \centering
    \includegraphics[width=1.0\columnwidth]{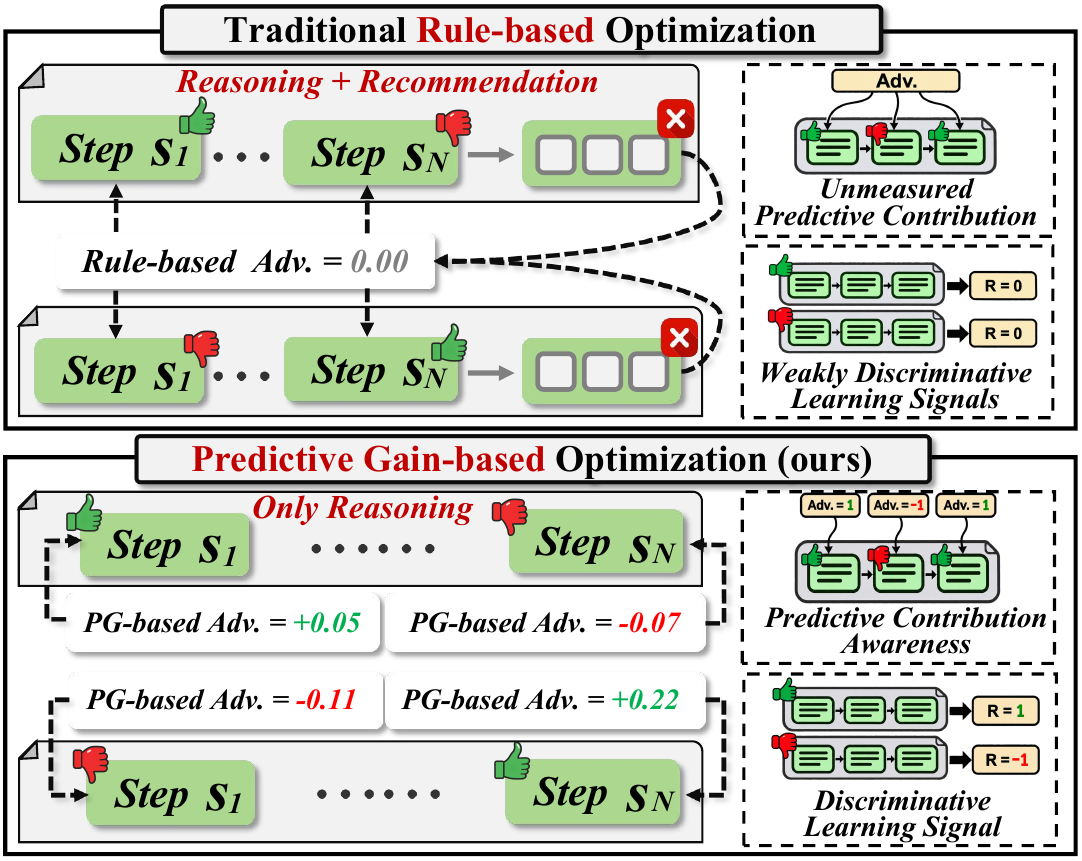}
    \caption{Comparison between rule-based and predictive gain (PG)-based reasoning optimization. The former relies on the performance of recommendations after reasoning, without measuring the predictive contributions within reasoning or adequately distinguishing rewards across reasoning trajectories. In contrast, PG evaluates the predictive contributions of individual reasoning sentences, providing more predictive contribution-aware and discriminative learning signals.}
    \label{fig:intro_case}
\end{figure}

Generative recommendation (GR) has emerged as a promising
paradigm for sequential recommendation, particularly through
Semantic ID (SID)-based item generation~\cite{
tiger,lladarec,rpg,lcrec,onerec}.
Inspired by the reasoning capabilities of large language
models (LLMs)~\cite{deepseek,cot1,cot2}, recent studies have
introduced explicit reasoning into GR to infer evolving
user interests and preferences for next-item
prediction~\cite{onerec-think,onereason,sidreasoner,evo-rec,gream}.
Compared with direct SID prediction, explicit reasoning
offers a promising way to leverage the world knowledge
encoded in LLMs to infer latent preferences beyond observed
interaction patterns, facilitating target SID prediction.

To realize this potential, existing methods typically
first fine-tune LLMs on chain-of-thought (CoT) trajectories
to establish reasoning behaviors, followed by 
reasoning optimization via Group Relative Policy
Optimization (GRPO)~\cite{deepseek_math}.
Such optimization is commonly guided by predefined rules,
such as SID matching~\cite{onerec-think,onereason,
openonerec,sidreasoner,gream,sapo}.
However, recent studies suggest that introducing explicit reasoning
does not consistently improve recommendation performance
and may even lead to degradation~\cite{
disconnect,why-think-hurts,onereason}, raising concerns
about the effectiveness of current optimization strategies.

We investigate existing rule-based reasoning optimization strategies and identify two key limitations.
\textbf{(i) Intra-Reasoning: Unmeasured Predictive Contributions.}
Before reasoning optimization, GR models typically acquire recommendation capabilities through extensive recommendation training~\cite{lcrec}. Thus, strong recommendation performance does not necessarily imply that reasoning improves recommendation. Yet existing rule-based methods primarily evaluate the performance of recommendations after reasoning without directly measuring the predictive contributions as reasoning progresses. Our empirical analysis further reveals nearly equal proportions of reasoning sentences with positive and negative predictive contributions during training, yet  rule-based optimization signals do not adequately reflect the contribution differences. 
\textbf{(ii) Intra-Group: Weakly Discriminative Learning Signals.}
Exactly predicting the target item from a large item space
is inherently challenging in GR~\cite{setrec,HRPO}. 
Consequently, reasoning trajectories with different
predictive contributions may receive identical rewards 
under predefined rules.
Although existing methods employ relaxed 
criteria~\cite{gream,sidreasoner,onerec-think},
their optimization signals may still fail to distinguish
these differences across rollouts.
Our empirical analysis confirms that all rewards are identical within a substantial proportion of rollout groups, which further leads to zero group-relative 
advantages and limits optimization efficiency.

To address these limitations, a natural approach is to employ external models
to evaluate intermediate reasoning~\cite{mgfrec,retrieval-grounded},
but assessing reasoning contributions to recommendation
remains costly and prone to bias. 
Inspired by gain-based agentic
RL~\cite{igpo,igpo2,igpo3,stepwise},
we introduce \textbf{Predictive Gain} (PG)
as the core reward signal, which quantifies changes in target SID log-likelihood
induced by reasoning steps. 
PG offers three desirable properties:
\textbf{(i) Predictive Contribution Awareness.}
PG explicitly measures the predictive
contributions of individual reasoning steps, revealing
their effects on recommendation throughout the
reasoning process.
\textbf{(ii) Discriminative Learning Signal.}
PG assigns continuous, fine-grained rewards to individual reasoning steps, enabling more discriminative feedback across trajectories with different predictive contributions. 
\textbf{(iii) Ranking Alignment.}
We theoretically establish that increasing expected PG tightens an upper bound on Top-$K$ ranking regret, providing a principled alignment between reasoning optimization and the recommendation objective.

Building on these insights, we propose \textbf{GainReason}, a simple yet effective training framework for reasoning-enhanced GR. Following prior works~\cite{onerec-think,onereason,sapo,sidreasoner,evo-rec,gream}, we first equip LLMs with recommendation capabilities through hybrid recommendation tasks and activate reasoning behaviors via fine-tuning on offline constructed CoT trajectories. We then introduce PG-based Reasoning Optimization (PGRO), which assigns sentence-level PG rewards to optimize reasoning generation using a GRPO-style surrogate objective. By quantifying the predictive contributions of individual reasoning sentences, PGRO enables fine-grained policy optimization without decoding recommendation outcomes or relying on predefined rules for reward computation. We also develop a custom causal attention mechanism to efficiently compute PG rewards across reasoning sentences. To complement PGRO, we further introduce an auxiliary PG-based Reasoning Utilization (PGRU) task that strengthens reasoning-conditioned target SID prediction for trajectories with near-zero net predictive gains. 

Our main contributions are summarized as follows:
\begin{itemize}[leftmargin=*, nosep]
    \item To the best of our knowledge, this work is the first to introduce predictive gain for optimizing explicit reasoning in GR. We investigate two limitations of existing rule-based optimization: unmeasured predictive contributions within reasoning and weakly discriminative learning signals within rollout groups. 

    \item We propose \textbf{GainReason}, a simple yet
    effective training framework that
    leverages PG to provide predictive contribution-aware and
    discriminative learning signals for reasoning optimization. 
    We further theoretically establish the alignment
    between expected PG and recommendation ranking.

    \item We conduct extensive experiments on public
    benchmarks, demonstrating the effectiveness of
GainReason against competitive reasoning-enhanced GR
methods. Further analyses examine the contributions of individual optimization components, training efficiency, and  reasoning behavior. 
\end{itemize}
 \section{Preliminaries}

\subsection{Task Formulation}

Let $\mathcal{I}$ denote the item set and
$(v_1,\ldots,v_T)$ a user's chronological interaction
history, where $v_t\in\mathcal{I}$.
Each item is represented by $L$ discrete Semantic ID
(SID) tokens obtained through quantization methods
such as RQ-VAE~\cite{tiger} and RQ-KMeans~\cite{onerec}.
Let $\mathbf{y}_t=\operatorname{SID}(v_t)$.
The SID-based history is
$H=[\mathbf{y}_1;\ldots;\mathbf{y}_T]$,
and the next-item target is
$\mathbf{y}^{*}=\operatorname{SID}(v_{T+1})$.
A standard generative recommender directly predicts
the target SID:
\begin{equation}
    \widehat{\mathbf{y}}
    \sim \pi_{\theta}(\cdot\mid H).
\end{equation}

\subsection{Learning to Reason}

Generative recommendation with explicit reasoning
typically follows three training stages~\cite{
onerec-think,onereason,openonerec,sidreasoner,
gream,evo-rec,sapo,retrieval-grounded}.

\subsubsection{\textbf{Itemic Alignment.}}

The vocabulary of a pretrained LLM $\pi_{\theta_0}$
is extended with SID tokens.
The model is then fine-tuned on tasks such as next-item
prediction and text--SID translation to integrate
recommendation knowledge with its pretrained
linguistic representations.
Given the resulting input--output pairs
$\mathcal{D}_{\mathrm{IA}}$, the objective is
\begin{equation}
    \mathcal{L}_{\mathrm{IA}}(\theta)
    =
    -\mathbb{E}_{(\mathbf{x},\mathbf{z})
    \sim\mathcal{D}_{\mathrm{IA}}}
    \left[
        \sum_{t=1}^{|\mathbf{z}|}
        \log\pi_{\theta}
        (z_t\mid\mathbf{x},\mathbf{z}_{<t})
    \right].
\end{equation}
This stage yields an aligned model $\pi_{\theta_1}$
with basic SID understanding and recommendation
capabilities.

\subsubsection{\textbf{Reasoning Activation.}}

Itemic alignment alone does not ensure stable explicit
reasoning generation~\cite{onerec-think,sidreasoner}.
Reasoning activation typically uses an offline CoT 
dataset, where an additional LLM generates reasoning
$\boldsymbol{O}^{\mathrm{T}}$ from textual descriptions
of the history and target item.
Each supervision sequence concatenates the generated
reasoning with the ground-truth SID:
$\mathbf{z}^{\mathrm{T}}
=[\boldsymbol{O}^{\mathrm{T}};\mathbf{y}^{*}]$.
The aligned model $\pi_{\theta_1}$ is then fine-tuned
on the resulting dataset $\mathcal{D}_{\mathrm{RA}}$:
\begin{equation}
    \mathcal{L}_{\mathrm{RA}}(\theta)
    =
    -\mathbb{E}_{(H,\mathbf{z}^{\mathrm{T}})
    \sim\mathcal{D}_{\mathrm{RA}}}
    \left[
        \sum_{t=1}^{|\mathbf{z}^{\mathrm{T}}|}
        \log\pi_{\theta}
        (z_t^{\mathrm{T}}\mid
        H,\mathbf{z}_{<t}^{\mathrm{T}})
    \right].
\end{equation}
This stage yields $\pi_{\theta_2}$, which generates
reasoning followed by an SID prediction conditioned
on the interaction history.

\subsubsection{\textbf{Reasoning Optimization.}}

The activated model $\pi_{\theta_2}$ is further
optimized through RL, typically using
GRPO~\cite{deepseek_math}.
For each history $H$, the old policy
$\pi_{\theta_{\mathrm{old}}}$ samples $G$ complete
outputs
$\mathbf{z}_i=
[\boldsymbol{O}_i;\widehat{\mathbf{y}}_i]$,
where $\boldsymbol{O}_i$ denotes the reasoning.
Each output receives a rule-based reward
$R_i^{\mathrm{SID}}
=\mathcal{R}_{\mathrm{SID}}
(\widehat{\mathbf{y}}_i,\mathbf{y}^{*})$.
Group-wise normalization gives
\begin{equation}
    \widetilde{A}_i
    =
    \frac{R_i^{\mathrm{SID}}-\mu_{\mathrm{SID}}}
         {\sigma_{\mathrm{SID}}+\epsilon},
\end{equation}
where $\mu_{\mathrm{SID}}$ and $\sigma_{\mathrm{SID}}$
are the group reward mean and standard deviation,
and $\epsilon>0$ ensures numerical stability.
The policy minimizes
\begin{equation}
\begin{aligned}
    \mathcal{L}_{\mathrm{RO}}(\theta)
    ={}&
    -\mathbb{E}\left[
        \frac{1}{G}\sum_{i=1}^{G}
        \frac{1}{|\mathbf{z}_i|}
        \sum_{t=1}^{|\mathbf{z}_i|}
        \operatorname{CLIP}_{\eta}
        \big(\rho_{i,t}(\theta),\widetilde{A}_i\big)
    \right]
    \\
    &+\beta\,\mathcal{D}_{\mathrm{KL}}
    \big(\pi_{\theta}\Vert\pi_{\mathrm{ref}}\big),
    \label{eq:grpo}
\end{aligned}
\end{equation}
where
$\rho_{i,t}(\theta)=
\frac{
    \pi_{\theta}(z_{i,t}\mid H,\mathbf{z}_{i,<t})
}{
    \pi_{\theta_{\mathrm{old}}}
    (z_{i,t}\mid H,\mathbf{z}_{i,<t})
}$is the importance ratio. 

\begin{figure}[t]
    \centering
    \includegraphics[width=1.0\columnwidth]{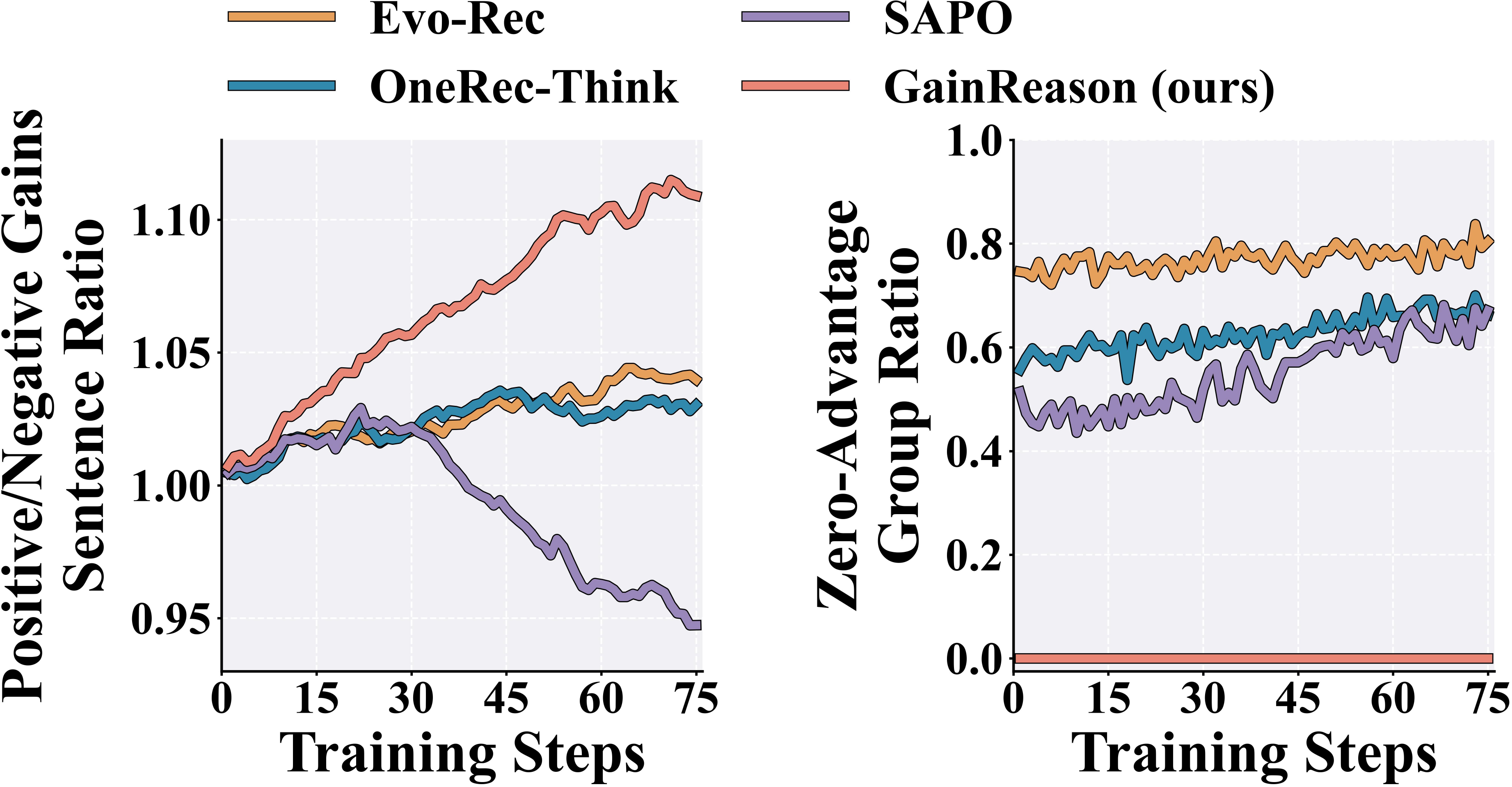}
    \caption{Analysis of reasoning optimization during training on Industrial. (Left) Ratio of reasoning sentences with positive to negative gains in target SID likelihood. (Right) Ratio of rollout groups with all zero-advantages. Compared with rule-based methods, GainReason achieves a higher ratio of positive to negative predictive gains and provides more discriminative learning signals, with no zero-advantage groups observed during training.}
    \label{fig:preliminary}
\end{figure}

\begin{figure*}[t]
    \centering
    \includegraphics[width=1.0\textwidth]{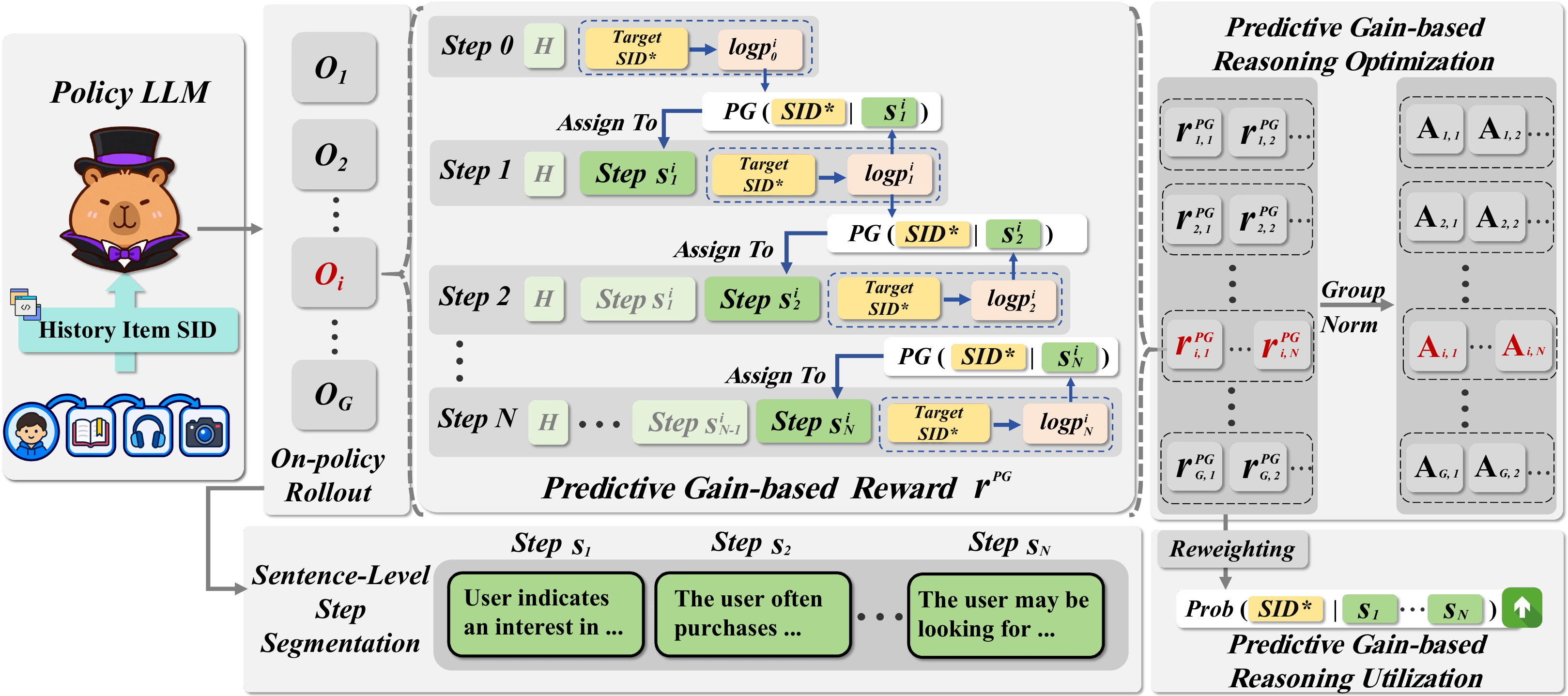} 
\caption{Overview of GainReason. PG-based Reasoning Optimization (PGRO) assigns sentence-level incremental predictive gains as rewards to optimize reasoning generation through group-relative policy optimization. An auxiliary PG-based Reasoning Utilization (PGRU) task adaptively enhances reasoning-conditioned target SID prediction by emphasizing trajectories with near-zero net predictive gains. Itemic alignment and reasoning activation are omitted for clarity.}

    \label{fig:methods}
\end{figure*}

\section{Methodology}

\subsection{Motivation}
Existing reasoning-enhanced generative recommenders~\cite{
onereason,onerec-think,gream,sidreasoner,evo-rec,sapo}
typically rely on rule-based optimization,
where rewards depend primarily on recommendation performance.
However, this paradigm faces two key limitations:

\noindent\textbf{(i) Intra-Reasoning: Unmeasured Predictive Contributions.}
While reasoning aims to improve target SID prediction,
rule-based methods primarily evaluate final recommendation
outcomes without directly measuring the predictive
contributions of reasoning progress.
Our empirical analysis (Fig.~\ref{fig:preliminary}(left))
reveals comparable proportions of reasoning sentences
with positive and negative gains in target SID likelihood
during rule-based training.
However, these distinct contributions are not adequately
reflected in existing optimization signals, leaving
their predictive effects insufficiently captured
during reasoning optimization.

\noindent\textbf{(ii) Intra-Group: Weakly Discriminative Learning Signals.}
Predicting the target item from a large item space is inherently challenging~\cite{setrec,HRPO}, making rule-based rewards sparse and insufficiently discriminative across rollouts. Consequently, entire rollout groups may receive zero group-relative advantages, limiting effective policy updates. As shown in Fig.~\ref{fig:preliminary}(right), Evo-Rec, which employs NDCG-based rewards, exhibits a particularly high proportion of zero-advantage groups. Although existing methods relax matching criteria, such as OneRec-Think's maximum matching score among beam candidates, or provide process-level feedback, such as SAPO's assignment of SID matching-based rewards to individual reasoning blocks, their learning signals may still fail to distinguish reasoning trajectories with different predictive contributions, resulting in relatively high zero-advantage group ratios. 

These limitations motivate us to introduce predictive
gain-based rewards that directly quantify the predictive
contributions of reasoning progress to target SID prediction,
providing contribution-aware and discriminative
learning signals.

\subsection{Model Preparation}
\textbf{SID construction.} Unless specified, we adopt RQ-KMeans~\cite{onerec}
to tokenize item textual metadata into discrete SIDs,
which is widely used in industrial generative recommendation
for its efficiency and strong performance. We extend the vocabulary of a pretrained LLM $\pi_{\theta_0}$, such as Qwen3~\cite{qwen3}, with the resulting SID tokens. 

\textbf{Itemic Alignment.} Following prior works~\cite{onerec-think, openonerec, onereason, sidreasoner, evo-rec}, we perform SFT with two categories of itemic alignment tasks:  (1) Item prediction. Given a user's interaction history, the model predicts the next item, with historical and target items represented as SID sequences or textual descriptions. This task captures behavioral patterns and connects SID representations with item semantics.   
(2) Item-SID translation. The model learns bidirectional mappings between item titles and their corresponding SID sequences, generating a title from an SID and vice versa. This task explicitly aligns SID tokens with item semantics in the LLM's linguistic space.

\textbf{Reasoning Activation.} Starting from the aligned model $\pi_{\theta_1}$, we perform reasoning activation through CoT SFT on offline-constructed reasoning trajectories. Following the CoT construction principles of OneReason~\cite{onereason}, each trajectory summarizes user behavior and preferences, explores plausible interest directions, and weighs behavioral evidence to infer the next purchase. The resulting model $\pi_{\theta_2}$ serves as the initial policy for subsequent RL optimization. Details are provided in Appendix~\ref{sec: training_details}.
\vspace{-5pt}

\subsection{Predictive Gain-based Reward}
We view explicit reasoning as \textit{an incremental process of uncovering information in users' interactions that facilitates the next item prediction}. To capture this intuition, we introduce an intrinsic \textit{predictive gain (PG)-based reward}, as detailed below. 

\textbf{Reasoning Step Segmentation.}
Fine-grained process rewards require identifiable step 
boundaries, such as tool-call markers in search
agents~\cite{igpo,igpo2,igpo3}.
In generative recommendation, however, explicit reasoning is generated as free-form text, which typically lacks explicit step delimiters. 
Since sentences serve as natural, semantically meaningful
units of text~\cite{sentence_segmentation},
we segment a reasoning rollout $\boldsymbol{O}$
using sentence-ending punctuation marks:
\begin{equation}
    \operatorname{Split}(\boldsymbol{O};\mathcal{P})
    =
    (s_1,\ldots,s_N),
\end{equation}
where $\mathcal{P}$ denotes the set of sentence-ending
punctuation marks, such as periods,
$s_j$ represents the $j$-th reasoning step,
and $N$ is the total number of steps.
This segmentation allows us to evaluate changes
in target SID prediction as the reasoning prefix
expands from
$\boldsymbol{O}_{<j}=[s_1;\ldots;s_{j-1}]$
to
$\boldsymbol{O}_{\leq j}=[s_1;\ldots;s_j]$,
providing the basis for sentence-level reward assignment.

\textbf{Predictive Gain Reward.}
Formally, let $\mathbf{y}^{*}=(y_1^{*},\ldots,y_L^{*})$
denote the ground-truth target SID.
For the $j$-th reasoning step, we compute the
log-likelihood of $\mathbf{y}^{*}$ under the
current policy $\pi_{\theta}$ as
\begin{equation}
    \ell_j
    =
    \frac{1}{L}\sum_{l=1}^{L}
    \log\pi_{\theta}
    \big(
        y_l^{*}\mid
        H,\boldsymbol{O}_{\leq j},
        \mathbf{y}_{<l}^{*}
    \big),
    \label{eq:sid_prob}
\end{equation}
where $\boldsymbol{O}_{\leq j}=[s_1;\ldots;s_j]$
denotes the reasoning prefix through step $j$.
The immediate PG reward for this step is defined as
\begin{equation}
    r_j^{\mathrm{PG}}
    =
    \ell_j-\ell_{j-1},
    \qquad j=1,\ldots,N,
    \label{eq:step_ig}
\end{equation}
where $\ell_0$ is computed using only the
interaction history $H$.

This sentence-level PG reward offers several desirable
properties.
(i) Predictive Contribution Awareness.
PG directly measures the predictive
contribution of each reasoning sentence to target SID
prediction, without relying on external reward 
models~\cite{mgfrec,retrieval-grounded} or manually
defined process-level rules~\cite{sapo}.
(ii) Discriminative Learning Signals.
PG provides continuous and fine-grained feedback,
enabling the differentiation of reasoning trajectories
with distinct predictive contributions. 
Beyond these advantages, however, PG is used to
optimize intermediate reasoning generation rather than
final recommendations, raising the question of whether
such optimization aligns with the ranking objective of GR. 
We theoretically establish a third property:
(iii) Ranking Alignment.
Specifically, we show that maximizing expected PG
given the current reasoning state is equivalent to
minimizing a ranking-consistent recommendation
surrogate, thereby tightening an upper bound on 
Top-$K$ ranking regret:
\begin{equation}
\begin{aligned}
&\mathbb{E}_{s_j\sim\pi_\theta(\cdot\mid H,\boldsymbol{O}_{<j})}
[\mathcal{R}_K(p_j)]
\\
&\quad\leq
\sqrt{
2\left[
D_{\mathrm{KL}}(\eta_H\Vert p_{j-1})
-
\mathbb{E}_{\substack{
s_j\sim\pi_\theta(\cdot\mid H,\boldsymbol{O}_{<j})\\
y^*\sim\eta_H
}}
\left[L\,r_j^{\mathrm{PG}}\right]
\right]
}.
\end{aligned}
\label{eq:ig_ranking_bound}
\end{equation} 
where $\eta_H$ denotes the true next-item distribution,
$p_j(v)=\pi_\theta(v\mid H,\boldsymbol{O}_{\leq j})$
is the predicted item distribution, and
$\mathcal{R}_K$ denotes Top-$K$ ranking regret. This result further provides a principled justification
for leveraging PG as a ranking-aligned reward signal 
for reasoning optimization in GR. 
The detailed proof is provided in
Appendix~\ref{sec:ranking_alignment}.

\begin{figure}[t]
    \centering
    \includegraphics[width=1.0\columnwidth]{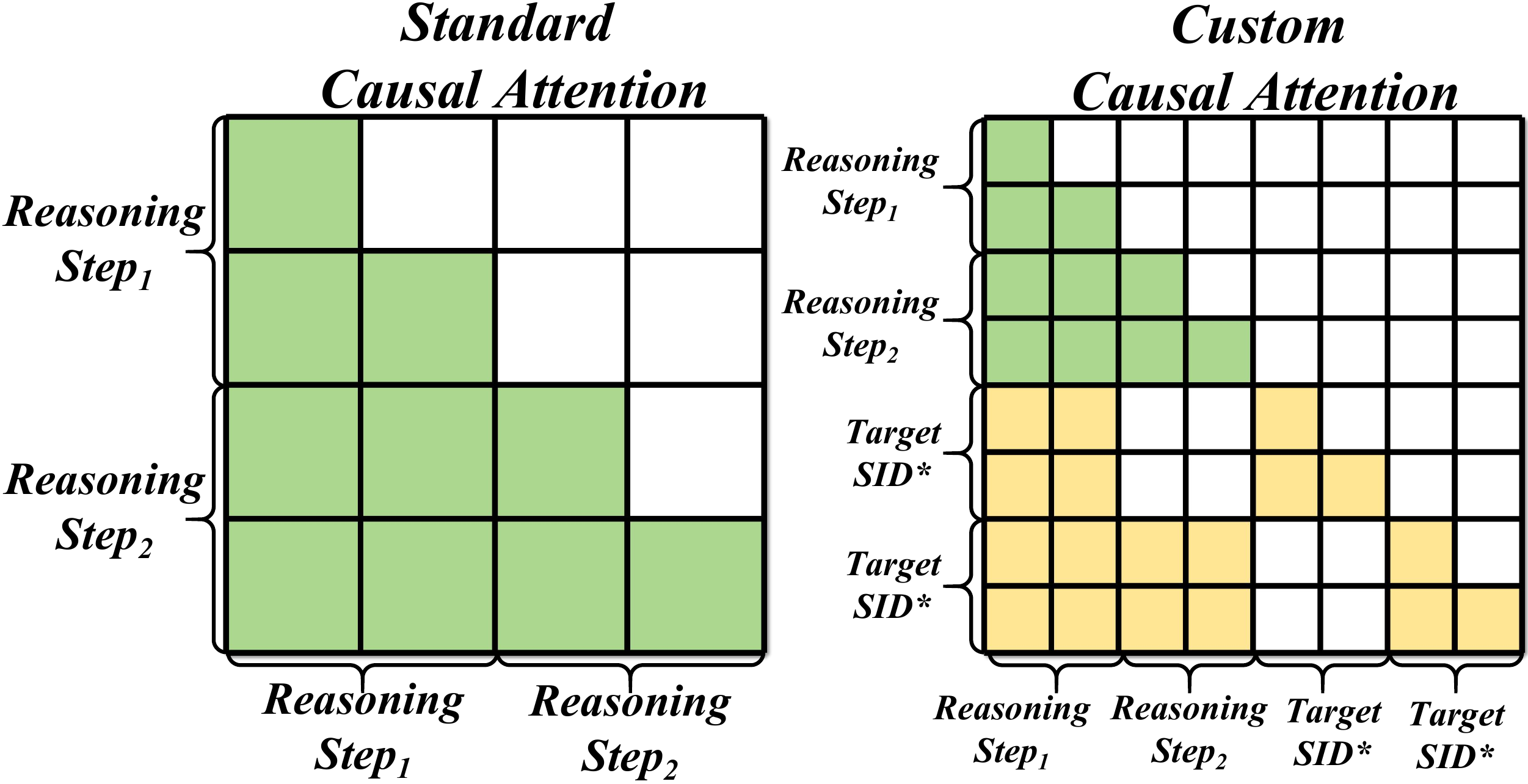}
    \caption{The custom causal attention for efficient PG computation. Shaded cells denote allowed attention. Prompt tokens and user history are omitted for clarity.}
    \label{fig:attention}
\end{figure}

\textbf{Efficient PG Computation.}
Naively computing PG rewards requires a separate
forward pass for each reasoning prefix, incurring
substantial computational overhead due to repeated
processing of the shared user history and overlapping
reasoning prefixes.
To address this issue, we develop a custom causal
attention mechanism (Fig.~\ref{fig:attention}) for
efficient PG reward computation.
Specifically, for a reasoning rollout
$\boldsymbol{O}=[s_1;\ldots;s_N]$ consisting of
$N$ segmented reasoning sentences, we append
$N+1$ copies of the target SID to the
complete reasoning sequence, including one for
the history-only baseline.
Through the custom causal attention mask, each
copy attends only to the interaction history $H$,
its corresponding reasoning prefix, and its own
preceding SID tokens.
With position IDs matched to those of separate
prefix evaluations, all prefix scores
$\{\ell_j\}_{j=0}^{N}$ can be computed in a
single forward pass.
The PG rewards are then obtained from adjacent
score differences (Eq.~\ref{eq:step_ig}),
avoiding repeated computation over shared prefixes. 

\subsection{Policy Optimization}

\textbf{PG-based Reasoning Optimization (PGRO).} 
For each training instance $(H,\mathbf{y}^*)\sim\mathcal{D}$,
we sample $G$ reasoning rollouts
$\mathcal{O}=\{\boldsymbol{O}_i\}_{i=1}^{G}$ from the old policy
$\pi_{\theta_{\mathrm{old}}}$.
Each rollout $\boldsymbol{O}_i$ consists of $N_i$
reasoning sentences, with corresponding PG rewards \footnote{We apply
stop-gradient to the PG rewards during optimization.}
$\{r_{i,j}^{\mathrm{PG}}\}_{j=1}^{N_i}$
computed using Eq.~\ref{eq:step_ig}. 
Following GRPO, we adopt group-wise z-normalization
to stabilize policy optimization and capture the relative
magnitudes of sentence-level predictive contributions:
\begin{equation}
    A_{i,j}
    =
    \frac{r^{\mathrm{PG}}_{i,j}-\mu_{\mathrm{PG}}}
         {\sigma_{\mathrm{PG}}},
    \qquad 1\leq j\leq N_i,
    \label{eq:on_policy_advantage}
\end{equation}
where $\mu_{\mathrm{PG}}$ and $\sigma_{\mathrm{PG}}$
denote the mean and standard deviation of all
sentence-level PG rewards within the rollout group.
Each reasoning token inherits the advantage of its
corresponding sentence, i.e., $A_{i,t}=A_{i,j}$
for token $t$ belonging to sentence $s_j^i$.

We formulate the reasoning optimization objective as:

\begin{equation}
\mathcal{L}_{\mathrm{PGRO}}(\theta)
=
-\mathbb{E}_{(H,\mathbf{y}^*),\mathcal{O}}
\left[
\frac{1}{G}
\sum_{i=1}^{G}
\frac{1}{|\boldsymbol{O}_i|}
\sum_{t=1}^{|\boldsymbol{O}_i|}
\operatorname{CLIP}_{\eta}
\left(
\rho_{i,t}(\theta),
A_{i,t}
\right)
\right], 
\label{eq:igro}
\end{equation}
where $\operatorname{CLIP}_{\eta}(\rho,A)=\min\big(\rho A,\operatorname{clip}(\rho,1-\eta,1+\eta)A\big)$ denotes the clipped GRPO surrogate, $\rho_{i,t}(\theta)
=\frac{\pi_{\theta}(o_{i,t}\mid H,o_{i,<t})}
{\pi_{\theta_{\mathrm{old}}}(o_{i,t}\mid H,o_{i,<t})}$
is the importance ratio. 


\textbf{PG-based Reasoning Utilization (PGRU).}
While PGRO optimizes reasoning generation based on its
predictive contributions, near-zero predictive gains may
arise not only from ineffective reasoning but also from
insufficient utilization of generated reasoning
~\cite{making_reasoning_matter,measuring_faithful}.
To complement PGRO, we introduce an auxiliary
PG-based Reasoning Utilization
(PGRU) task to strengthen reasoning-conditioned
prediction, particularly for trajectories with
near-zero predictive gains. 

Specifically, we aggregate sentence-level PG rewards to obtain the trajectory-level PG reward, $R_i^{\mathrm{PG}}=\sum_{j=1}^{N_i}r_{i,j}^{\mathrm{PG}}$. We prioritize trajectories with small-magnitude $R_i^{\mathrm{PG}}$ for auxiliary supervision, as large positive values indicate that reasoning has already yielded substantial predictive gains, while large negative values may reflect misleading reasoning that additional supervision could reinforce. In contrast, small-magnitude $R_i^{\mathrm{PG}}$ indicates limited net predictive gains, potentially arising from either insufficient reasoning quality, which PGRO seeks to improve, or ineffective reasoning utilization. Such trajectories are therefore suitable for additional reasoning-conditioned supervision with less potential interference with PGRO.  
Motivated by these considerations, we employ a smooth bump 
function~\cite{Tu2011}
to assign PG-adaptive utilization weights: 
\begin{equation}
w_i =
\begin{cases}
\displaystyle
\exp\left(-\frac{u_i^2}{1-u_i^2}\right),
& |u_i|<1, \\[6pt]
0, & |u_i|\geq 1,
\end{cases}
\quad
u_i=\frac{2R_i^{\mathrm{PG}}}{W},
\label{eq:igru_weight}
\end{equation}
where $W>0$ controls the effective range of utilization
enhancement. The weighting function peaks at zero PG,
decreases smoothly as the magnitude of PG increases,
and vanishes when $|R_i^{\mathrm{PG}}|\geq \frac{W}{2}$.  We then incorporate these PG-adaptive weights into
reasoning-conditioned fine-tuning:
\begin{equation}
\mathcal{L}_{\mathrm{PGRU}}(\theta)
=
-\mathbb{E}_{(H,\mathbf{y}^*)\sim\mathcal{D},\,\mathcal{O}}
\left[
\frac{1}{G}\sum_{i=1}^{G}
\operatorname{sg}(w_i)
\log\pi_\theta
\left(\mathbf{y}^*\mid H,\boldsymbol{O}_i\right)
\right],
\label{eq:igru_objective}
\end{equation}
where the generated reasoning trajectories are
treated as fixed conditioning contexts.
In this way, PGRU adaptively enhances reasoning utilization
for trajectories with near-zero net PG.

Finally, GainReason integrates PGRO and PGRU through
the following joint objective:
\begin{equation}
\mathcal{L}_{\mathrm{GainReason}}(\theta)
=
\mathcal{L}_{\mathrm{PGRO}}(\theta)
+ 
\lambda\mathcal{L}_{\mathrm{PGRU}}(\theta),
\label{eq:IG-Rec_objective}
\end{equation}
where $\lambda\geq 0$ balances the contribution of PGRU.

\subsection{Discussion}
\subsubsection{\textbf{Comparison with Rule-based Methods}}
The essential distinction between GainReason and existing
rule-based reasoning optimization methods lies in
how optimization rewards are defined.
Existing methods primarily derive rewards from
the performance of recommendations after reasoning.
Specifically, OneRec-Think~\cite{onerec-think} uses the maximum
SID matching score among beam candidates.
SIDReasoner~\cite{sidreasoner} combines prefix-based
matching rewards with format rewards.
Evo-Rec~\cite{evo-rec} evaluates the target item's
ranking position within beam-search results using NDCG.
SAPO~\cite{sapo} further assigns SID  matching
rewards to individual reasoning blocks.
In contrast, GainReason directly measures the
predictive contributions of reasoning sentences
through changes in target SID log-likelihood.
This distinction offers three key advantages for
policy optimization:
\begin{itemize}[leftmargin=*, nosep]
    \item\textbf{Reward Density.}
Rule-based rewards are typically sparse and
often identical across rollouts within a group,
resulting in zero group-relative advantages.
In contrast, GainReason provides continuous and
discriminative rewards, enabling more rollout groups
to contribute effective learning signals
for policy optimization.

    \item \textbf{Credit Assignment Granularity.}
    GainReason evaluates the predictive contributions
    of individual reasoning sentences, allowing different
    sentences within the same trajectory to receive
    distinct advantages based on their relative contributions.
    This enables more finer-grained credit assignment
    for reasoning optimization.

    \item \textbf{Decoding-Free Reward Computation.} 
    GainReason directly measures changes in target SID
    likelihood as reasoning progresses, without requiring
    explicit recommendation decoding or handcrafted 
     rules.
    This simplifies reward design and reduces
    engineering complexity.
\end{itemize}

By shifting the reward basis from final recommendation
performance to the predictive gains of reasoning 
progress, GainReason provides more informative and
objective-aligned learning signals and streamlined
policy optimization.

\subsubsection{\textbf{Training Computational Budget}}
GainReason follows the standard GRPO training pipeline,
with additional forward computation primarily arising
from sentence-level PG estimation.
Our custom causal attention mechanism avoids repeated
evaluation of reasoning prefixes by extending the
original sequence from $s=t+\ell_{\mathrm{SID}}$
to $t+(N+1)\ell_{\mathrm{SID}}$, where $t$,
$\ell_{\mathrm{SID}}$, and $N$ denote the text length,
target SID length, and number of reasoning sentences,
respectively.
Under the conditions detailed in
Appendix~\ref{sec:flops_analysis},
the relative FLOPs overhead $\delta$ of the
extended forward pass is bounded by:
\begin{equation}
\delta =
\frac{\Delta F}{F_{\mathrm{GRPO}}}
< \frac{2N_{\max}\ell_{\mathrm{SID}}}{s},
\end{equation}
where $N_{\max}$ denotes the maximum number of
reasoning sentences.
This establishes an explicit upper bound on the
additional computational cost and reveals a natural
trade-off between computational overhead and
reasoning granularity: finer segmentation requires
more PG evaluations but enables more precise reward
assignment based on the predictive contributions
of reasoning.
The detailed derivation is provided in
Appendix~\ref{sec:flops_analysis}, with empirical
evaluation of training efficiency presented in
Section~\ref{sec:compute_analysis}.
\begin{table}[t]
    \centering
    \caption{Statistics of the benchmark datasets.}
    \label{tab:dataset_statistics}
    \small
    \setlength{\tabcolsep}{3pt}
    \renewcommand{\arraystretch}{1.1}
    \resizebox{\columnwidth}{!}{%
        \begin{tabular}{@{}lrrrrr@{}}
            \toprule
            Dataset & \#Users & \#Items & \#Inter.
                    & Avg.\ Len. & Sparsity \\
            \midrule
            Industrial & 7,694  & 3,686   & 49,371
                       & 3.98  & 99.8259\% \\
            Movies     & 14,216 & 13,190  & 148,099
                       & 6.01 & 99.9210\% \\
            Yelp       & 19,940 & 15,390  & 211,844
                       & 6.14 & 99.9310\% \\
            \bottomrule
        \end{tabular}%
    }
\end{table}
\section{Experiments}
\subsection{Experimental Setup}
\subsubsection{\textbf{Datasets.}}
We evaluate our method on three widely used benchmarks: \textbf{Industrial \& Scientific}, and \textbf{Movies \& TV} from the Amazon review corpus~\cite{amazon}, and the \textbf{Yelp} business review dataset. We apply 5-core filtering to retain users and items with at least five interactions. To prevent temporal leakage across data splits~\cite{leakage}, we globally sort all interactions across users by their absolute timestamps and split them into training, validation, and test sets in an 8:1:1 ratio. Following prior works~\cite{sidreasoner,evo-rec}, we construct each user's history from preceding interactions using a sliding window of up to 10 items. For SID construction, we encode item metadata with Sentence-T5 and quantize the resulting embeddings using RQ-KMeans~\cite{onerec}, representing each item as a sequence of three discrete SID tokens. The metadata includes the title, brand, category, and description for Amazon items, and the name, category, city, and state for Yelp businesses. Table~\ref{tab:dataset_statistics} summarizes the processed dataset statistics.

\begin{table*}[t]
\centering
\footnotesize
\renewcommand{\arraystretch}{1.2}
\setlength{\tabcolsep}{3.0pt}
\caption{Overall performance on three recommendation benchmarks with different categories of methods. The best and second-best results are highlighted in bold and \underline{underline}, respectively.}
\label{tab:performance_mean}

\resizebox{\textwidth}{!}{%
\begin{tabular}{l*{12}{c}}
\toprule
\multirow{2}{*}{\textbf{Models}}
& \multicolumn{4}{c}{\textbf{Industrial}}
& \multicolumn{4}{c}{\textbf{Movies}}
& \multicolumn{4}{c}{\textbf{Yelp}} \\
\cmidrule(lr){2-5}
\cmidrule(lr){6-9}
\cmidrule(lr){10-13}
& N@5 & N@10 & HR@5 & HR@10
& N@5 & N@10 & HR@5 & HR@10
& N@5 & N@10 & HR@5 & HR@10 \\
\midrule

\multicolumn{13}{c}{%
    \cellcolor{gray!15}\footnotesize
    \textit{Traditional Sequential Recommenders}} \\
\midrule
GRU4Rec
& 0.0541 & 0.0580 & 0.0673 & 0.0797
& 0.0077 & 0.0088 & 0.0107 & 0.0141
& 0.0138 & 0.0195 & 0.0226 & 0.0393 \\
BERT4Rec
& 0.0609 & 0.0642 & 0.0767 & 0.0872
&0.0071	&0.0094	&0.0111	&0.0180
& 0.0150 & 0.0211 & 0.0244 & 0.0434 \\
SASRec
& 0.0727 & 0.0776 & 0.0992 & 0.1145
&0.0118 &0.0150 &0.0187 &0.0289
& 0.0157 & 0.0219 & 0.0244 & 0.0439 \\
\midrule

\multicolumn{13}{c}{%
    \cellcolor{gray!15}\footnotesize
    \textit{Generative Recommenders}} \\
\midrule
TIGER
& 0.0762 & 0.0854 & 0.0990 & 0.1279
& 0.0124 & 0.0164 & 0.0200 & 0.0322
& 0.0173 & 0.0238 & 0.0277 & 0.0482 \\
RPG
& 0.0683 & 0.0771 & 0.0897 & 0.1175
& 0.0183 & 0.0222 & 0.0248 & 0.0370
& 0.0160 & 0.0211 & 0.0264 & 0.0435 \\
LLaDA-Rec
& 0.0787 & 0.0896 & 0.1012 & 0.1349
&0.0122 &0.0156	&0.0194&0.0301 
& 0.0170 & 0.0231 & 0.0279 & 0.0481 \\
LC-Rec
& 0.0871 & 0.0974 & 0.1085 & 0.1426
& 0.0409 & 0.0443 & 0.0518 & 0.0631
& 0.0179 & 0.0237 & 0.0284 & 0.0453 \\
\midrule

\multicolumn{13}{c}{%
    \cellcolor{gray!15}\footnotesize
    \textit{Reasoning-Enhanced Generative Recommenders}} \\
\midrule
ReaRec
& 0.0832 & 0.0909 & 0.1064 & 0.1300
& 0.0418 & 0.0457 & 0.0505 & 0.0624
& 0.0170 & 0.0237 & 0.0263 & 0.0472 \\
CARE
& 0.0758 & 0.0873 & 0.0960 & 0.1331
& 0.0326 & 0.0384 & 0.0434 & 0.0613
& 0.0164 & 0.0231 & 0.0266 & 0.0478 \\
OneRec-Think
& \underline{0.0915} & \underline{0.1028} & 0.1138 & \underline{0.1494}
& 0.0415 & 0.0460 & 0.0515 & 0.0652
& \underline{0.0191} & \underline{0.0254} & 0.0302 & \underline{0.0497} \\
SIDReasoner
& 0.0903 & 0.1023 & \underline{0.1149} & 0.1493
& 0.0420 & \underline{0.0466} & 0.0525 & \underline{0.0666}
& 0.0187 & 0.0242 & 0.0295 & 0.0466 \\
Evo-Rec
& 0.0902 & 0.1009 & 0.1132 & 0.1464
& \underline{0.0422} & 0.0464 & \underline{0.0526} & 0.0657
& 0.0188 & 0.0249 & 0.0293 & 0.0483 \\
SAPO
& 0.0878 & 0.0997 & 0.1089 & 0.1458
&0.0413 &0.0461 &0.0522 &0.0664
&0.0187 &0.0241 &\underline{0.0303} &0.0470 \\
\midrule

\rowcolor{blue!10}
\textbf{GainReason (ours)}
& \textbf{0.0985} & \textbf{0.1092} & \textbf{0.1227} & \textbf{0.1557}
& \textbf{0.0435} & \textbf{0.0479} & \textbf{0.0541} & \textbf{0.0680}
& \textbf{0.0204} & \textbf{0.0270} & \textbf{0.0320} & \textbf{0.0526} \\
\bottomrule
\end{tabular}%
}
\end{table*}

\subsubsection{\textbf{Baseline Methods.}}
We compare GainReason with the following categories of baselines:
\begin{itemize}[leftmargin=*, nosep]
    \item \textbf{Traditional Sequential Recommenders:}
    GRU4Rec~\cite{gru4rec}, SASRec~\cite{sasrec}, and BERT4Rec~\cite{bert4rec}.

    \item \textbf{Generative Recommenders:}
    TIGER~\cite{tiger}, RPG~\cite{rpg}, LLaDA-Rec~\cite{lladarec}, and LC-Rec~\cite{lcrec}.

    \item \textbf{Reasoning-Enhanced Generative Recommenders:}
    We compare our method with two types of reasoning strategies: 
    
    (1)\textit{Latent Reasoning:}
        ReaRec~\cite{rearec} and CARE~\cite{care}. (2) \textit{Explicit Reasoning:}
        OneRec-Think~\cite{onerec-think},
        SIDReasoner~\cite{sidreasoner},  Evo-Rec~\cite{evo-rec}, and SAPO~\cite{sapo}.
        To ensure a fair comparison of RL optimization strategies, all  explicit reasoning-based methods are initialized from the same checkpoint.
\end{itemize}
Detailed descriptions of the baselines are provided in Appendix~\ref{sec:baselines}.

\subsubsection{\textbf{Evaluation Metrics.}}
Following prior works~\cite{sidreasoner,evo-rec}, we evaluate recommendation performance using Hit Ratio (HR@$K$) and Normalized Discounted Cumulative Gain (NDCG@$K$), with $K\in\{5,10\}$. HR@$K$ measures whether the target item appears among the top-$K$ recommendations, while NDCG@$K$ additionally accounts for its ranking position. All evaluations use the full item catalog rather than sampled candidates to avoid sampling bias. 

\subsubsection{\textbf{Implementation details.}} Please refer to Appendix~\ref{sec:implementation} for detailed implementation and hyperparameter settings.

\subsection{Overall Performance}
The overall performance comparison is reported in Table~\ref{tab:performance_mean}. We summarize several key observations as follows:

\begin{itemize}[leftmargin=*, nosep]
\item Among different recommendation paradigms, explicit reasoning methods generally outperform traditional sequential recommenders, conventional GR methods, and latent reasoning approaches. For instance, on Industrial, OneRec-Think achieves an NDCG@5 of 0.0915, surpassing SASRec (0.0727), LC-Rec (0.0871), and ReaRec (0.0832). Similarly, Evo-Rec achieves an NDCG@5 of 0.0422 on Movies, outperforming LC-Rec (0.0409) and the latent reasoning methods ReaRec (0.0418) and CARE (0.0326). These results demonstrate the merits of explicit reasoning in leveraging the world knowledge encoded in LLMs to infer user preferences and improve recommendation performance.

\item Existing rule-based reasoning optimization methods exhibit comparable performance despite adopting different reward designs. For example, OneRec-Think, SIDReasoner, and Evo-Rec achieve closely clustered NDCG@5 scores of 0.0915, 0.0903, and 0.0902 on Industrial, respectively, with similar patterns observed on Movies and Yelp. These methods primarily compute rewards based on final recommendation performance and assign them to entire reasoning trajectories, without directly measuring the predictive contributions of individual reasoning sentences. SAPO further introduces process-level feedback by associating SID matching-based rewards with corresponding reasoning blocks. However, such heuristic assignments may not faithfully reflect the actual predictive contributions of reasoning, potentially limiting their effectiveness. Empirically, SAPO performs comparably to trajectory-level approaches rather than consistently outperforming them, suggesting that finer-grained matching feedback alone may be insufficient for effective reasoning optimization.

\item GainReason achieves superior performance across various datasets, with average relative improvements of 6.2\%, 2.7\%, and 6.1\% over the strongest baselines on Industrial, Movies, and Yelp, respectively. Specifically, on Industrial, GainReason improves NDCG@5 by 7.7\%, HR@5 by 6.8\%, and HR@10 by 4.2\%. On Movies, it achieves improvements ranging from 2.1\% to 3.1\% across the four metrics, while on Yelp, the improvements reach 6.8\% for NDCG@5, 6.3\% for NDCG@10, 5.6\% for HR@5, and 5.8\% for HR@10. These consistent improvements demonstrate the effectiveness of GainReason. Unlike existing rule-based methods, GainReason directly measures the  predictive contributions of reasoning and provides fine-grained, discriminative supervision without relying on final recommendation performance, enabling more effective optimization of reasoning for next-item prediction.
\end{itemize}

\subsection{Reasoning Effectiveness Analysis}

\begin{figure}[t]
    \centering
    \includegraphics[width=1.0\columnwidth]{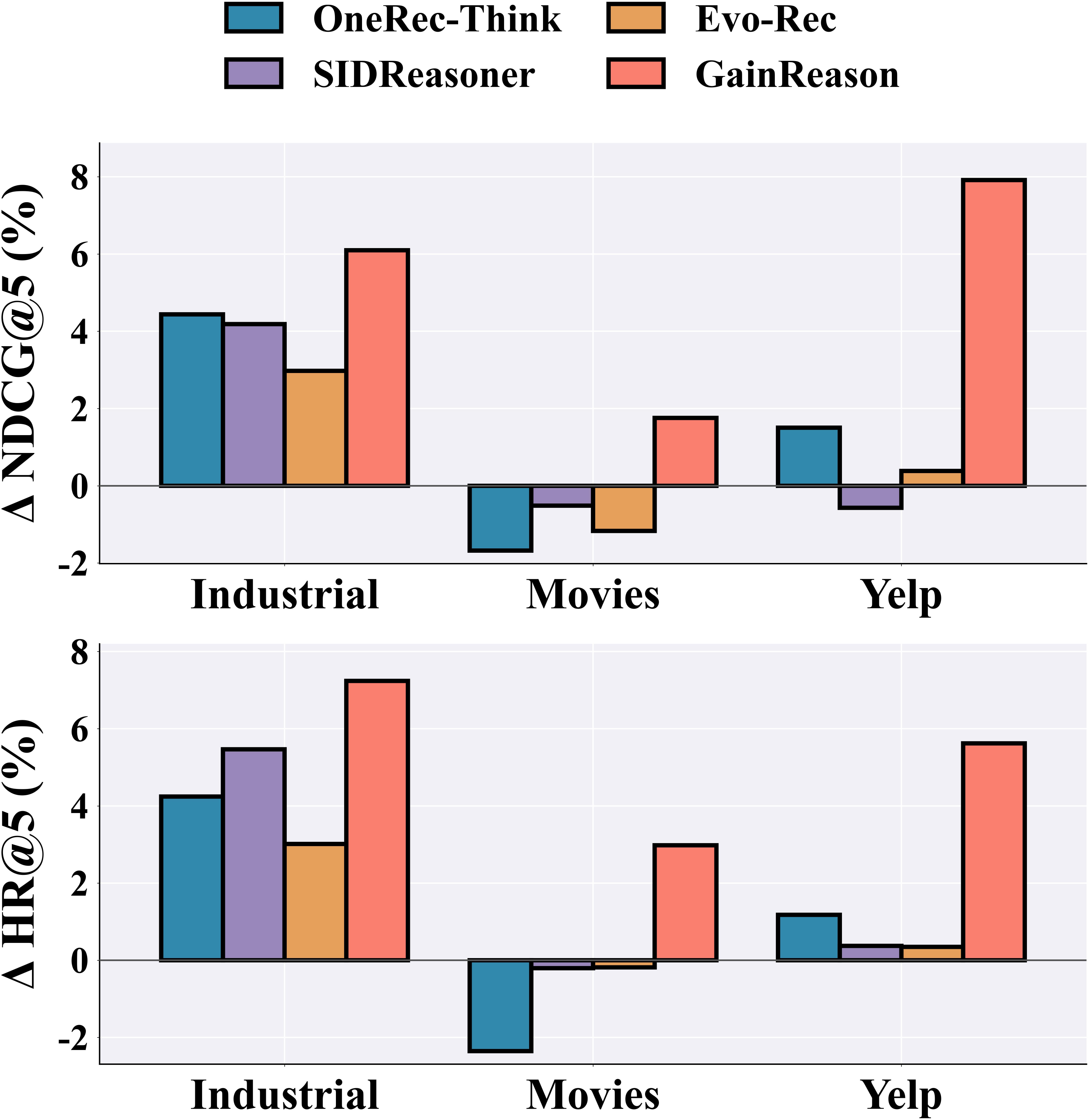}
    \caption{Changes in recommendation performance from enabling
    reasoning across methods and datasets.}
    \label{fig:reasoning_effectiveness}
\end{figure}

To examine whether explicit reasoning genuinely benefits
recommendation, we compare the performance of each method
with and without reasoning, as illustrated in
Fig.~\ref{fig:reasoning_effectiveness}.
Three key observations emerge as follows:

\textbf{(i)} Existing rule-based methods exhibit inconsistent improvements from explicit reasoning across datasets. While reasoning generally improves recommendation performance on Industrial, it leads to performance degradation on Movies and yields relatively limited improvements on Yelp. These findings are consistent with previous studies showing that explicit reasoning does not necessarily improve recommendation performance and may even impair it~\cite{disconnect,why-think-hurts,onereason}.

\textbf{(ii)} In contrast, GainReason consistently benefits
from explicit reasoning across all evaluated settings.
Enabling reasoning yields average relative improvements
of 5.8\%, 2.3\%, and 7.4\% on Industrial, Movies,
and Yelp, respectively, exceeding those of existing
methods.
Notably, even on Movies, where rule-based methods
frequently experience performance degradation,
GainReason consistently achieves positive improvements.
On Yelp, it further exhibits substantially greater
benefits from reasoning than the compared methods.
These results suggest that PG-based optimization
more effectively translates explicit reasoning into
recommendation improvements. 

\textbf{(iii)} Further comparison with
Table~\ref{tab:performance_mean} reveals that GainReason's advantages on Movies and
Yelp stem primarily from the additional
improvements brought by reasoning, instead of stronger direct recommendation
capabilities. 
On Movies, GainReason performs slightly worse than
Evo-Rec without reasoning on HR@5
(0.0525 vs. 0.0528), but surpasses it after
reasoning (0.0541 vs. 0.0526).
Similarly, on Yelp, GainReason only marginally
outperforms OneRec-Think in direct recommendation
on HR@5 (0.0303 vs. 0.0299), yet further widens
its lead after reasoning (0.0320 vs. 0.0302).
These results highlight the substantial contribution
of reasoning to GainReason's performance advantages
in these settings, demonstrating the effectiveness
of optimizing the predictive contributions of reasoning.

\subsection{In-Depth Analysis}
\subsubsection{\textbf{Ablation Study}}
We conduct ablation studies on Industrial and Yelp to evaluate the contributions of different optimization components. Specifically, we remove PGRO (w/o PGRO) and PGRU (w/o PGRU) to examine the effects of reasoning optimization and utilization, respectively. We also remove PG-adaptive weighting (w/o PG-adaptive Weighting) to evaluate its benefit over uniform reasoning-conditioned prediction supervision.

The ablation results are presented in Table~\ref{tab:ablation}, from which we make three observations. (1) Removing PGRO leads to a substantial performance decline, indicating that reasoning-conditioned prediction supervision alone cannot achieve the improvements obtained by the full framework. This highlights the importance of  optimizing reasoning based on its predictive contributions. (2) PGRO alone already outperforms existing baselines, demonstrating the effectiveness of PG-based reasoning optimization. Incorporating PGRU brings further improvements, suggesting that enhancing reasoning utilization complements reasoning optimization. (3) Replacing PG-adaptive weighting with uniform supervision yields inferior results, indicating that selectively strengthening prediction for trajectories with near-zero net predictive gains is more effective than uniformly supervising all trajectories.

\begin{table}[t]
    \caption{Ablation results on Industrial and Yelp.
    ``w/o'' denotes removing a component. 
    }
    \label{tab:ablation}
    \centering
    \footnotesize
    \renewcommand{\arraystretch}{1.2}
    \setlength{\tabcolsep}{4pt}
    \setlength{\dashlinedash}{3pt}
    \setlength{\dashlinegap}{3pt}

    \resizebox{\columnwidth}{!}{%
    \begin{tabular}{@{}lcccc@{}}
        \toprule
        Model
        & \multicolumn{2}{c}{Industrial}
        & \multicolumn{2}{c}{Yelp} \\
        \cmidrule(lr){2-3}
        \cmidrule(lr){4-5}
        & N@5 & HR@5 & N@5 & HR@5 \\
        \midrule

        \textbf{GainReason (Ours)}
        &\textbf{0.0985} &\textbf{0.1227} &\textbf{0.0204} &\textbf{0.0320}\\
        \midrule

        After Preparation
        &0.0881 &0.1109 &0.0169 &0.0273 \\
        \midrule
        w/o PGRO
        &0.0892 &0.1125 &0.0191 &0.0304 \\
        w/o PGRU
        &0.0954 &0.1202 &0.0200 & 0.0315\\ 
        w/o PG-adaptive Weighting
        &0.0963 &0.1211 &0.0201 &0.0315 \\
        \bottomrule
    \end{tabular}%
    }
\end{table}

\subsubsection{\textbf{Training Dynamics.}}
\begin{figure}[t]
    \centering
    \includegraphics[width=1.0\columnwidth]{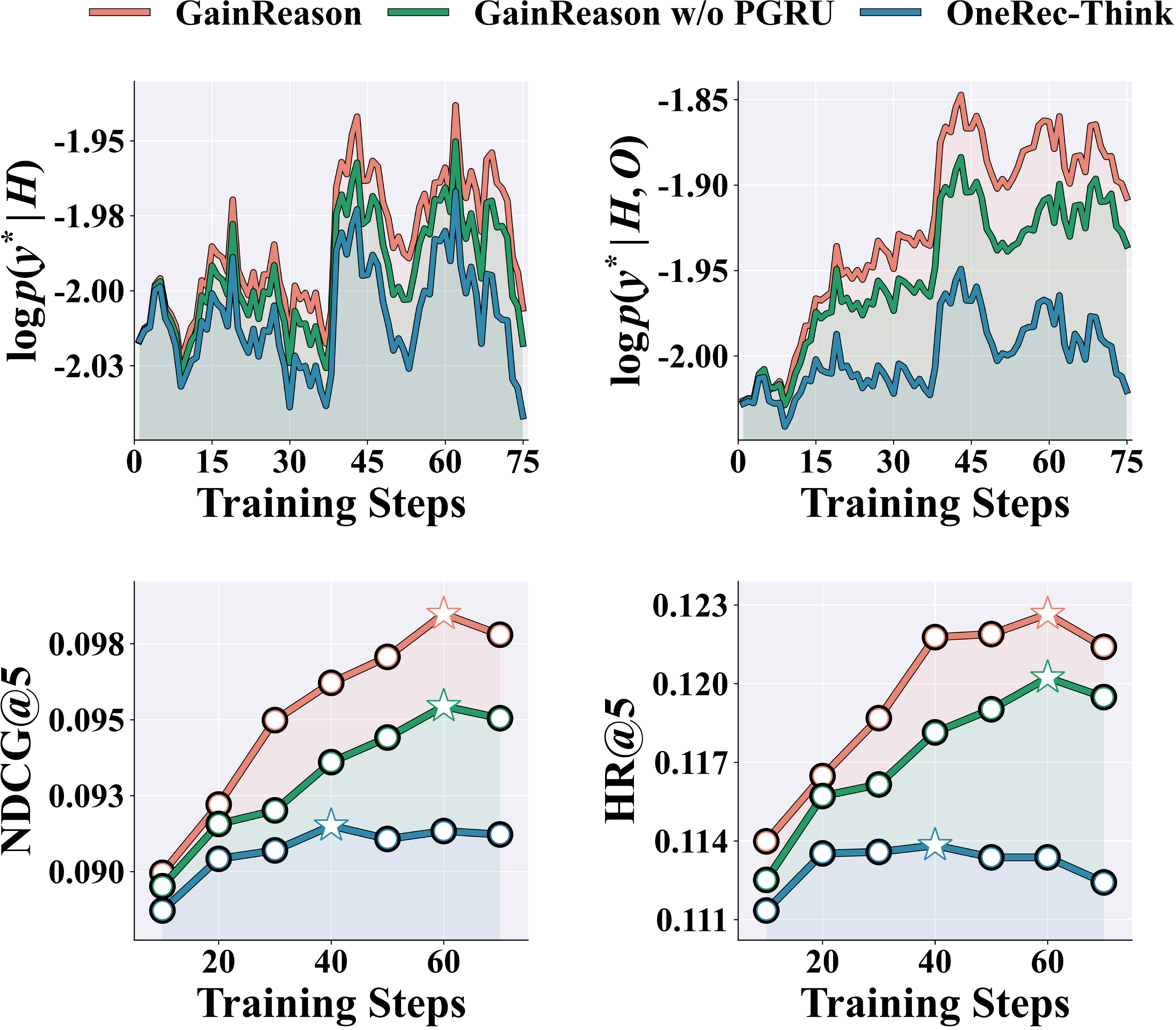}
    \caption{Target SID log-likelihood before and after
reasoning and recommendation performance dynamics
during reasoning optimization on Industrial.}
    \label{fig:training_dynamics}
\end{figure}
To investigate how PG-based reasoning optimization improves recommendation performance, we track the target SID log-likelihood with and without reasoning, alongside NDCG@5 and HR@5 throughout training. As shown in Fig.~\ref{fig:training_dynamics}, GainReason achieves substantially higher reasoning-conditioned target likelihood than OneRec-Think, while the differences in history-only likelihood remain relatively modest. This indicates that PG-based process supervision effectively enhances the predictive contributions of reasoning. Moreover, the improvements in reasoning-conditioned likelihood are accompanied by consistent gains in recommendation performance, demonstrating the benefits of optimizing reasoning based on its predictive contributions. Notably, GainReason without PGRU already outperforms OneRec-Think, while incorporating PGRU further improves both target likelihood and recommendation performance, highlighting the complementary role of reasoning utilization enhancement.

\subsubsection{\textbf{Training Efficiency Analysis}} 
\begin{figure}[t]
    \centering
    \includegraphics[width=1.0\columnwidth]{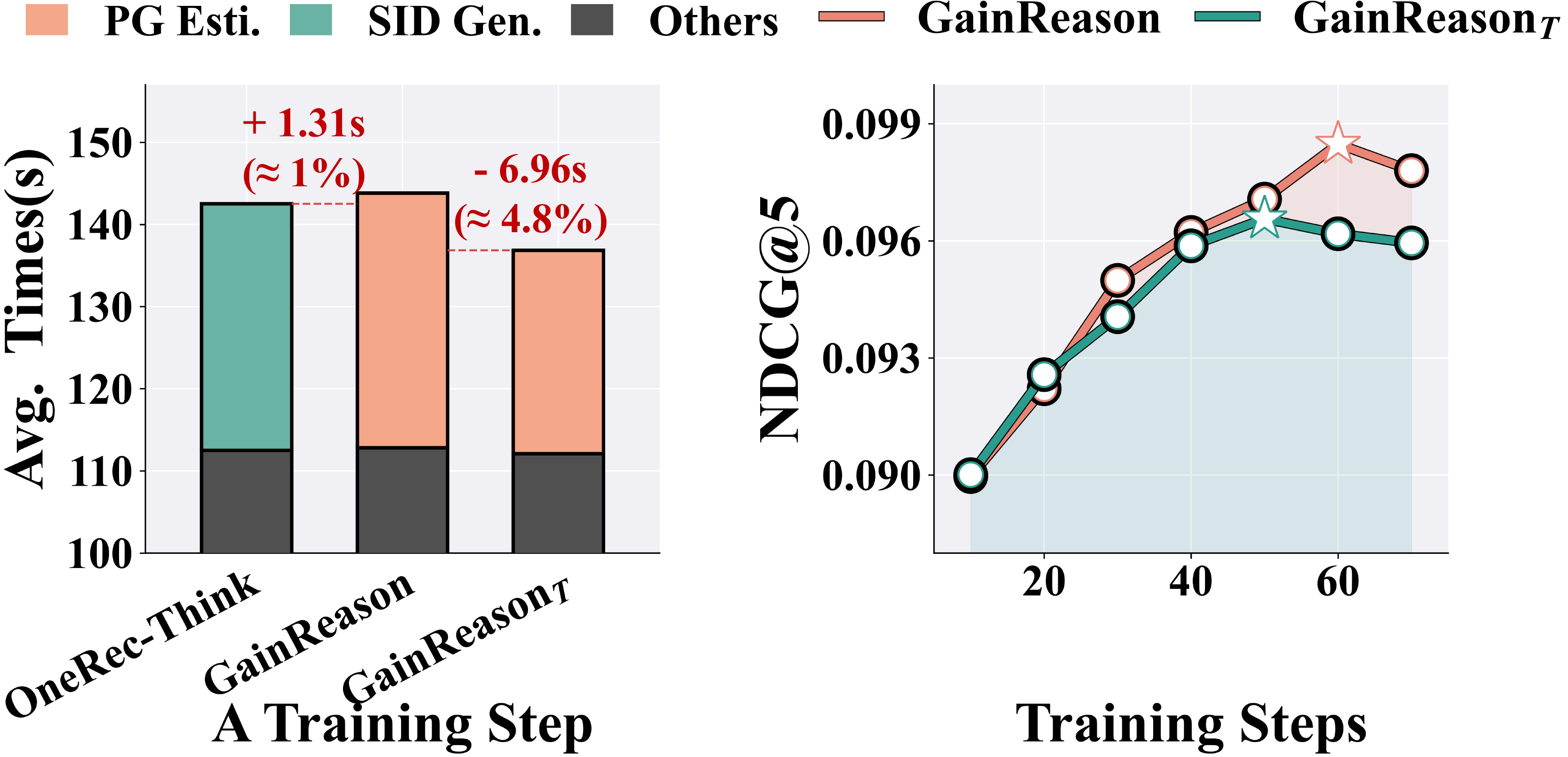}
    \caption{Training time per step for OneRec-Think,
    GainReason, and GainReason$_{\mathrm{T}}$ (left),
    and recommendation performance of the two GainReason variants
    on Industrial (right). 
    GainReason and GainReason$_{\mathrm{T}}$ use
    sentence-level and trajectory-level PG rewards, respectively.}
    \label{fig:training_computation}
\end{figure}

To evaluate the training efficiency of GainReason and the effectiveness of fine-grained PG assignment, we compare its training costs and optimization performance with OneRec-Think and a trajectory-level variant, GainReason$_T$, as shown in Fig.~\ref{fig:training_computation}. Despite requiring fine-grained reward estimation, GainReason increases the average training time per step by only 1.31 seconds (approximately 1\%) compared with OneRec-Think. This modest overhead reflects two counteracting factors. On the one hand, sentence-level PG estimation introduces additional costs from sentence segmentation, sequence preparation, and reward computation. Our custom causal attention mechanism specifically reduces the reward computation overhead by evaluating multiple reasoning prefixes in a single forward pass. On the other hand, GainReason avoids expensive beam search decoding for recommendation outcomes by directly computing target SID likelihood, thereby reducing decoding costs. Compared with GainReason, GainReason$_T$ further reduces the training time by 6.96 seconds (approximately 4.8\%) through trajectory-level reward estimation.

Despite this additional computational cost, GainReason achieves stronger recommendation performance than GainReason$_T$. While both variants measure changes in target SID log-likelihood induced by reasoning, they differ in the granularity of reward computation and assignment. Specifically, GainReason computes incremental PG between consecutive reasoning prefixes and assigns sentence-level rewards, whereas GainReason$_T$ computes the overall PG between history-only and complete reasoning predictions and assigns a uniform reward to the entire trajectory. By distinguishing the incremental predictive contributions within reasoning, sentence-level PG assignment provides more informative supervision for policy optimization. These results demonstrate that fine-grained PG assignment improves reasoning optimization at a modest additional computational cost.

\label{sec:compute_analysis}

\subsubsection{\textbf{Case Study.}}

\begin{figure}[t]
    \centering
    \includegraphics[width=1.0\columnwidth]{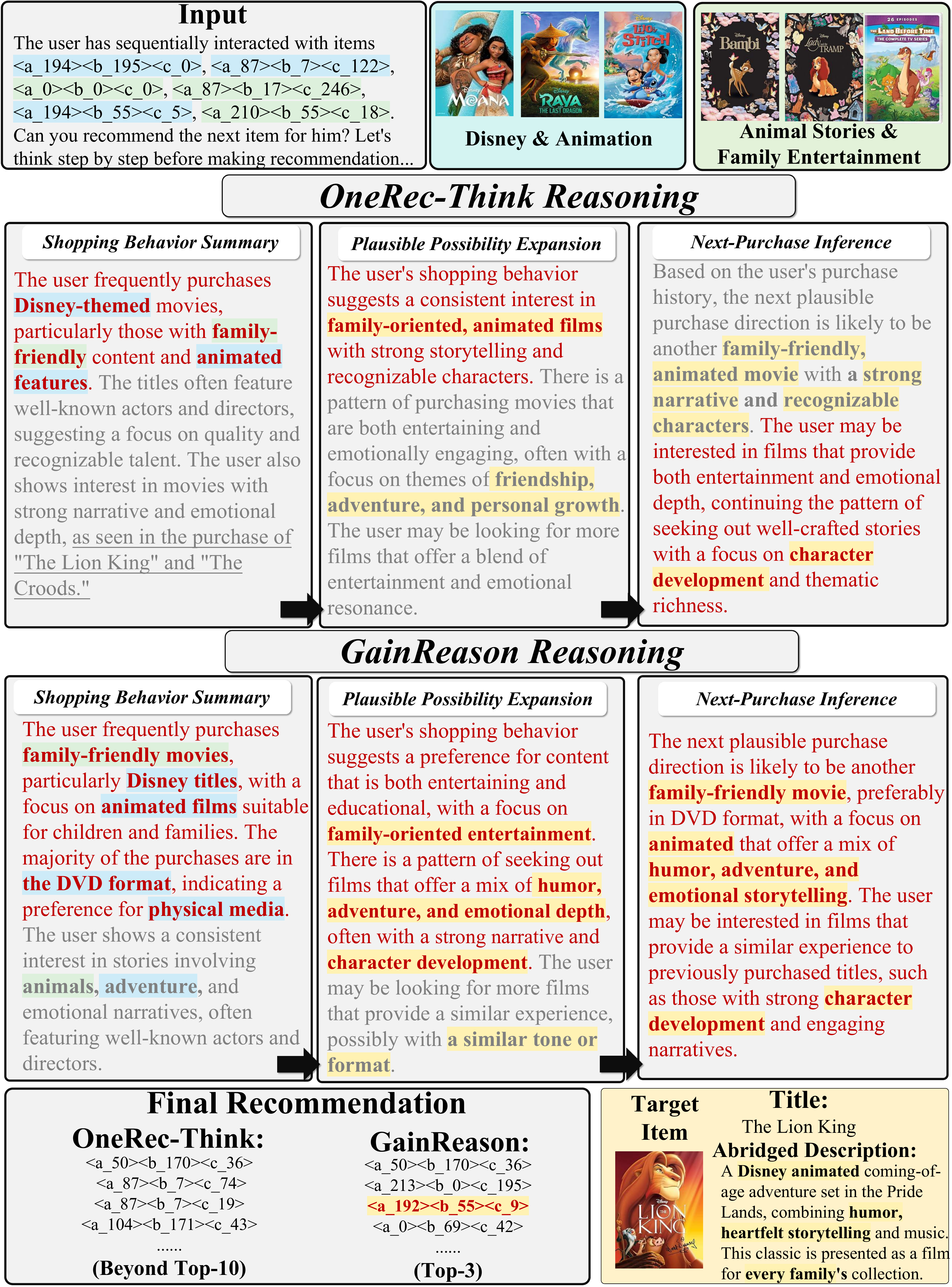}
    \caption{Case study comparing the reasoning behaviors
    of OneRec-Think and GainReason. 
    Sentences with positive and negative PG are shown in
    \textcolor[HTML]{C00000}{red} and \textcolor[HTML]{808080}{gray}, respectively. Shaded text highlights attributes related to the
    user history or target item, while \underline{underlined text} 
    indicates incorrect statements.}
    \label{fig:case_study}
\end{figure}

Figure~\ref{fig:case_study} presents a representative case comparing the reasoning behaviors of OneRec-Think and GainReason. Overall, GainReason produces more reasoning sentences with positive predictive gains than OneRec-Think (6 vs. 3). Further examination of their reasoning reveals that positive gains are often associated with accurate identification of relevant item attributes. Specifically, both methods correctly recognize \textit{``Disney-related''} and \textit{``family-friendly animations''} in the user history and infer preferences related to the target item, such as \textit{``adventure''}. However, OneRec-Think also misidentifies historical items and even incorrectly treats the target item as part of the interaction history, resulting in negative predictive gains. Moreover, some of its generic or insufficiently grounded statements are associated with negative gains. These findings suggest that OneRec-Think includes more vague or unsupported statements, whereas GainReason produces more specific and history-grounded reasoning, yielding more positive predictive gains. These observations are consistent with previous findings on the importance of grounding reasoning in recommendation~\cite{onereason}. In terms of recommendation outcomes, GainReason ranks the target item within the Top-3 recommendations, whereas OneRec-Think fails to rank it within the Top-10, illustrating how more effective reasoning can translate into improved recommendation ranking. 
Interestingly, we also observe that correctly grounded reasoning does not always yield positive predictive gains, particularly for OneRec-Think. This suggests that accurate preference inference alone is insufficient to guarantee predictive improvements, as useful reasoning information may not be effectively translated into target item prediction. This observation further supports the complementary role of PGRU in enhancing reasoning utilization.

\section{Related Work}
\textbf{Reasoning-enhanced Generative Recommendation.}
Building on sequential recommendation~\cite{sasrec,bert4rec},
generative recommendation (GR) formulates next-item
prediction as item generation, with Semantic ID (SID)-based
methods gaining increasing attention~\cite{
tiger,rpg,lladarec,setrec,MiniOneRec,etegrec}.
Recent studies integrate SIDs into pretrained LLMs to
unify language understanding and item prediction~\cite{lcrec,onerec}.
Inspired by LLM reasoning advances~\cite{
cot1,cot2,deepseek,deepseek_math},
reasoning-enhanced GR follows two main paradigms.
Latent reasoning performs inference through hidden
states~\cite{rearec,onepiece} or latent tokens~\cite{
latentr3,implict-reason},
whereas explicit reasoning generates textual rationales
before recommendation, typically through CoT fine-tuning
followed by rule-based RL~\cite{sidreasoner,evo-rec,gream}.
The latter has also been deployed in industrial
recommendation systems~\cite{onerec-think,onereason},
yet recent studies find that explicit reasoning does not
consistently improve recommendation performance
over direct prediction~\cite{disconnect,why-think-hurts}.
To improve reasoning optimization, some methods
introduce finer-grained reward assignment, such as
SAPO~\cite{sapo}, which assigns SID matching-based
rewards to individual reasoning blocks.
Others employ external models to evaluate intermediate
reasoning~\cite{mgfrec,retrieval-grounded}.
However, these approaches do not directly measure
the predictive contributions of individual reasoning
steps, while matching-based rewards may provide
insufficient discrimination across rollouts,
limiting effective policy optimization.
GainReason addresses these limitations through
predictive gain-based rewards, providing
contribution-aware and discriminative learning signals.

\textbf{Policy Optimization with Process Rewards.}
Rule-based RL typically evaluates answer
correctness, providing limited feedback on individual
reasoning steps~\cite{deepseek_math,deepseek,dapo,Open-Reasoner-Zero}.
Process reward models (PRMs) introduce intermediate
supervision, initially focusing on step correctness~\cite{
lets_verify,math_shepherd}.
However, correctness does not necessarily reflect
predictive contributions toward the final objective.
To address this limitation,
\citet{Rewarding_progress} measures changes in future
success likelihood, while recent search agent 
methods explore information gains. 
For example, IGPO~\cite{igpo} measures changes in
the policy's likelihood of the correct answer,
TIPS~\cite{igpo2} uses a teacher-derived potential,
and IG-Search~\cite{igpo3} measures the gain in answer
confidence from retrieved documents relative to
random-document baselines. 
Similar developments have emerged in recommendation,
where previous methods obtain intermediate feedback
from external models, including process reward models,
simulated user agents, and frozen retrievers~\cite{
promise,mgfrec,retrieval-grounded}.
More closely related, OneReason~\cite{onereason}
analyzes target-likelihood changes along explicit CoT
but does not use them for policy optimization.
HiLaR~\cite{hilar} incorporates marginal target-likelihood
gains into RL for latent reasoning, but aggregates these
gains with final recommendation rewards to compute
trajectory-level advantages. 
In contrast, to the best of our knowledge,
GainReason is the first to directly employ sentence-level
predictive gains as intrinsic rewards for optimizing
explicit reasoning in GR, without requiring external
reward models.

\section{Conclusion}
In this paper, we investigate existing  rule-based reasoning optimization in generative recommendation and identify two key limitations: unmeasured predictive contributions within reasoning and weakly discriminative learning signals across trajectories. To address these limitations, we propose GainReason, a simple yet effective framework that optimizes reasoning based on its predictive contribution to target SID prediction. Specifically, PGRO assigns sentence-level predictive gains as rewards for fine-grained policy optimization, with a custom causal attention mechanism enabling efficient reward computation without explicit recommendation decoding. An auxiliary PGRU task further strengthens reasoning-conditioned prediction by selectively emphasizing trajectories with near-zero cumulative gains. We also establish a theoretical connection between expected PG and the recommendation ranking objective, providing a principled justification for PG-based reasoning optimization. Extensive experiments demonstrate consistent improvements over competitive baselines, while ablation studies, training dynamics, and case studies verify that GainReason more effectively translates reasoning into recommendation gains with modest training overhead. Our findings highlight the importance of evaluating reasoning through its actual predictive contributions, offering a promising direction for more effective reasoning optimization in generative recommendation.






\normalem
\begin{verbatim}

\end{verbatim}


\bibliographystyle{ACM-Reference-Format}
\bibliography{reference}

\appendix

\section{Theoretical Analysis}
\subsection{Ranking Alignment of PG Rewards}
\label{sec:ranking_alignment}

This section provides theoretical support for the
ranking alignment of the PG reward. We first
establish the relationship between SID negative
log-likelihood (NLL) and Top-$K$ ranking regret,
and then extend the analysis to stepwise and
complete reasoning trajectories.

\noindent\textbf{Notations.}
Given a user interaction history $H$, let $\mathcal V$
denote the valid item set, $\mathbf y^*$ denote the
target SID of length $L$, and
$\eta_H(v)=P(\mathbf y^*=\mathbf y(v)\mid H)$
denote the true next-item distribution, where
$\mathbf y(v)$ is the unique SID of item $v$.
We write $\mathbf y^*\sim\eta_H$ for the induced
target SID distribution.
Let $\boldsymbol O_i=(s_1^i,\ldots,s_{N_i}^i)$
denote the $i$-th reasoning trajectory.
The predicted item distribution after sentence $j$
is $p_j^i(v)=\pi_\theta(v\mid H,\boldsymbol O_{i,\le j})$,
where $\pi_\theta(v\mid\cdot)$ denotes the complete
SID generation probability and $p_0^i$ is the
history-only distribution.
Following Eq.~\ref{eq:step_ig}, the stepwise PG is
$r_{i,j}^{\mathrm{PG}}=
\frac{1}{L}[\log p_j^i(\mathbf y^*)
-\log p_{j-1}^i(\mathbf y^*)]$,
with cumulative PG
$R_i^{\mathrm{PG}}=\sum_{j=1}^{N_i}r_{i,j}^{\mathrm{PG}}$.

We consider valid-SID probabilities
satisfying $\sum_{v\in\mathcal V}p_j^i(v)=1$
and $\eta_H(v)>0\Rightarrow p_j^i(v)>0$,
as ensured by normalized constrained SID scoring.
On-policy reasoning generation does not access
the target SID, yielding
$\mathbf y^*\perp\boldsymbol O_i\mid H$.
Our analysis considers the rollout generation
and reward evaluation within a policy-optimization
iteration, during which both reasoning generation
and SID scoring use fixed model parameters $\theta$.
For a fixed preceding prefix $\boldsymbol O_{i,<j}$,
let $q$ denote a target-independent candidate
distribution over the next sentence $s_j^i$,
and define
$\mathcal J_{i,j}(q)=
\mathbb E_{s_j^i\sim q,\mathbf y^*\sim\eta_H}
[r_{i,j}^{\mathrm{PG}}]$.

\noindent\textbf{Remark.}
Under the above mild conditions, our analysis
characterizes the ranking alignment of expected
raw PG rewards through state-conditional policy
comparisons with a fixed SID scoring function. 
Practical optimization employs group-wise normalization,
token-level averaging, and ratio clipping, which
are beyond the scope of this analysis.

\noindent\textbf{Lemma A.1}
(Ranking Consistency and Regret Control of NLL).
Building on the strict propriety of logarithmic
scoring rules~\cite{Gneiting2007StrictlyPS}
and existing analyses of Top-$K$ surrogate
losses~\cite{lapin2016loss,yang2020consistency},
we establish the following result.
Consider the population SID NLL
$\mathcal C_{\eta_H}(p)=
\mathbb E_{\mathbf y^*\sim\eta_H}
[-\log p(\mathbf y^*)]$
and a general Top-$K$ ranking utility
$U_{\eta_H}(\rho)=
\sum_{k=1}^{K}w_k\eta_H(v_{(k)})$,
where $\rho=(v_{(1)},\ldots,v_{(K)})$
and $1\geq w_1\geq\cdots\geq w_K\geq0$.
Let $\rho_{\eta_H}^*$ and $\rho_p^*$
denote rankings induced by $\eta_H$ and $p$,
respectively, and define the ranking regret as
$\mathcal R_K(p)=
U_{\eta_H}(\rho_{\eta_H}^*)
-U_{\eta_H}(\rho_p^*)$.
Minimizing the population SID NLL over the full
probability simplex recovers the true item
distribution and induces a Bayes-optimal ranking.
Moreover, the corresponding ranking regret is
bounded by
\begin{equation}
\mathcal R_K(p)
\leq
\sqrt{
2\left[
\mathcal C_{\eta_H}(p)
-\mathcal C_{\eta_H}(\eta_H)
\right]
}.
\label{eq:theory_nll_regret_bound}
\end{equation}

\noindent\textbf{Proof.}
The strict propriety of log-loss
~\cite{Gneiting2007StrictlyPS} implies
$\mathcal C_{\eta_H}(p)
-\mathcal C_{\eta_H}(\eta_H)
=D_{\mathrm{KL}}(\eta_H\Vert p)\geq0$,
so the population NLL is minimized at
$p=\eta_H$.
Since the position weights are non-increasing,
the rearrangement inequality ensures that
$\rho_{\eta_H}^*$ maximizes the expected
ranking utility.
For any ranking $\rho$, define
$m_{\rho}(v)=w_k$ if $v=v_{(k)}$ and
$m_{\rho}(v)=0$ otherwise.
Since $0\leq m_{\rho}(v)\leq1$ and
$U_{\eta_H}(\rho)=
\mathbb E_{v\sim\eta_H}[m_{\rho}(v)]$,
the total variation bound for expectations
of functions in $[0,1]$ yields
$|U_{\eta_H}(\rho)-U_p(\rho)|
\leq\operatorname{TV}(\eta_H,p)$
for any ranking $\rho$.
Together with the optimality of $\rho_p^*$
under $p$, this yields
\begin{equation}
\begin{aligned}
\mathcal R_K(p)
&\leq
2\operatorname{TV}(\eta_H,p)\\
&\leq
\sqrt{2D_{\mathrm{KL}}(\eta_H\Vert p)}\\
&=
\sqrt{2\left[
\mathcal C_{\eta_H}(p)
-\mathcal C_{\eta_H}(\eta_H)
\right]},
\end{aligned}
\end{equation}
where the second inequality follows from
Pinsker's inequality, completing the proof.
\hfill$\square$

\noindent\textbf{Theorem A.2}
(Ranking Alignment of Step-wise PG).
Under the above setting, consider a fixed
preceding reasoning prefix $\boldsymbol O_{i,<j}$
and the candidate next-sentence distribution $q$.
Maximizing the expected PG $\mathcal J_{i,j}(q)$
over $q$ is equivalent to minimizing the expected
post-step SID NLL, a ranking-consistent surrogate
for Top-$K$ recommendation.
Furthermore, the expected Top-$K$ ranking regret
satisfies
\begin{equation}
\begin{aligned}
\mathbb E_{s_j^i\sim q}
[\mathcal R_K(p_j^i)]
\leq
\sqrt{2\Big[
D_{\mathrm{KL}}(\eta_H\Vert p_{j-1}^i)
-L\mathcal J_{i,j}(q)
\Big]}.
\end{aligned}
\label{eq:theory_step_bound}
\end{equation}

\noindent\textbf{Proof.}
We first establish the relationship between
expected PG and the recommendation surrogate.
By the definition of PG and the 
reasoning generation process, we have
\begin{equation}
\begin{aligned}
\mathcal J_{i,j}(q)
&=
\mathbb E_{s_j^i\sim q,\mathbf y^*\sim\eta_H}
\left[
\frac{1}{L}\bigl(
\log p_j^i(\mathbf y^*)
-\log p_{j-1}^i(\mathbf y^*)
\bigr)
\right]\\
&=
\frac{1}{L}\left[
\mathbb E_{s_j^i\sim q,\mathbf y^*\sim\eta_H}
[\log p_j^i(\mathbf y^*)]
-\mathbb E_{\mathbf y^*\sim\eta_H}
[\log p_{j-1}^i(\mathbf y^*)]
\right].
\end{aligned}
\label{eq:proof_step_expansion}
\end{equation}

Using the population NLL
$\mathcal C_{\eta_H}(p)$
defined in Lemma A.1,
Eq.~\ref{eq:proof_step_expansion} becomes
\begin{equation}
\mathcal J_{i,j}(q)
=
\frac{1}{L}\left[
\mathcal C_{\eta_H}(p_{j-1}^i)
-\mathbb E_{s_j^i\sim q}
\left[\mathcal C_{\eta_H}(p_j^i)\right]
\right].
\label{eq:proof_surrogate_difference}
\end{equation}
Since $\mathcal C_{\eta_H}(p_{j-1}^i)$
is independent of $q$, we have
\begin{equation}
\arg\max_q\mathcal J_{i,j}(q)
=
\arg\min_q
\mathbb E_{s_j^i\sim q}
\left[\mathcal C_{\eta_H}(p_j^i)\right].
\label{eq:proof_surrogate_equivalence}
\end{equation}
By Lemma A.1, the population SID NLL is a
ranking-consistent recommendation surrogate
when optimized over the full item probability
simplex. Hence, maximizing expected PG is
aligned with minimizing the expected post-step
recommendation surrogate.

Next, we derive the exact relationship between
PG and predictive KL divergence.
For a given reasoning sentence $s_j^i$,
taking the expectation over
$\mathbf y^*\sim\eta_H$ gives
\begin{equation}
\begin{aligned}
\mathbb E_{\mathbf y^*\sim\eta_H}
[r_{i,j}^{\mathrm{PG}}]
&=
\frac{1}{L}\sum_{v\in\mathcal V}
\eta_H(v)
\log\frac{p_j^i(v)}{p_{j-1}^i(v)}\\
&=
\frac{1}{L}\sum_{v\in\mathcal V}
\eta_H(v)
\left[
\log\frac{\eta_H(v)}{p_{j-1}^i(v)}
-\log\frac{\eta_H(v)}{p_j^i(v)}
\right]\\
&=
\frac{1}{L}\left[
D_{\mathrm{KL}}(\eta_H\Vert p_{j-1}^i)
-D_{\mathrm{KL}}(\eta_H\Vert p_j^i)
\right].
\end{aligned}
\label{eq:proof_step_kl}
\end{equation}

Taking the expectation over $s_j^i\sim q$
and noting that $p_{j-1}^i$ does not depend
on the current reasoning sentence,
we obtain
\begin{equation}
\mathcal J_{i,j}(q)
=
\frac{1}{L}\left[
D_{\mathrm{KL}}(\eta_H\Vert p_{j-1}^i)
-\mathbb E_{s_j^i\sim q}
\left[
D_{\mathrm{KL}}(\eta_H\Vert p_j^i)
\right]
\right].
\label{eq:proof_expected_step_kl}
\end{equation}
Thus, expected PG exactly characterizes
the reduction in predictive KL divergence
induced by the reasoning sentence.

Finally, we establish the ranking regret bound.
By Lemma A.1 and the NLL--KL decomposition,
for any predicted item distribution $p$,
we have
\begin{equation}
\mathcal R_K(p)
\leq
\sqrt{2D_{\mathrm{KL}}(\eta_H\Vert p)}.
\label{eq:proof_regret_kl}
\end{equation}
Applying this inequality to $p_j^i$
and taking the expectation over the generated
reasoning sentence yields
\begin{equation}
\begin{aligned}
\mathbb E_{s_j^i\sim q}
[\mathcal R_K(p_j^i)]
&\leq
\mathbb E_{s_j^i\sim q}
\left[
\sqrt{
2D_{\mathrm{KL}}(\eta_H\Vert p_j^i)
}
\right]\\
&\leq
\sqrt{
2\mathbb E_{s_j^i\sim q}
\left[
D_{\mathrm{KL}}(\eta_H\Vert p_j^i)
\right]
},
\end{aligned}
\label{eq:proof_step_jensen}
\end{equation}
where the second inequality follows from
Jensen's inequality.
Substituting
Eq.~\ref{eq:proof_expected_step_kl}
into Eq.~\ref{eq:proof_step_jensen},
we obtain
\begin{equation}
\begin{aligned}
\mathbb E_{s_j^i\sim q}
[\mathcal R_K(p_j^i)]
&\leq
\sqrt{
2\mathbb E_{s_j^i\sim q}
\left[
D_{\mathrm{KL}}(\eta_H\Vert p_j^i)
\right]
}\\
&=
\sqrt{2\Big[
D_{\mathrm{KL}}(\eta_H\Vert p_{j-1}^i)
-L\mathcal J_{i,j}(q)
\Big]}.
\end{aligned}
\label{eq:proof_step_final}
\end{equation}
This establishes
Eq.~\ref{eq:theory_step_bound}
and completes the proof.
\hfill$\square$

\noindent\textbf{Corollary A.3}
(Ranking Alignment of Trajectory-wise PG).
Under the same setting, consider a complete
reasoning trajectory $\boldsymbol O_i$
containing $N_i$ sentences, where $N_i$
may vary across rollouts.
The expected cumulative PG characterizes
the reduction in predictive KL divergence
between the history-only and final
reasoning-conditioned item distributions.
Consequently, the expected Top-$K$
ranking regret after the complete reasoning
trajectory satisfies
\begin{equation}
\begin{aligned}
\mathbb E_{\boldsymbol O_i}
[\mathcal R_K(p_{N_i}^i)]
\leq
\sqrt{
2\Big[
D_{\mathrm{KL}}(\eta_H\Vert p_0^i)
-\mathbb E_{\boldsymbol O_i,\mathbf y^*}
[L R_i^{\mathrm{PG}}]
\Big]
}.
\end{aligned}
\label{eq:theory_cumulative_bound}
\end{equation}

\noindent\textbf{Proof.}
Using the stepwise PG--KL relationship
established in Eq.~\ref{eq:proof_step_kl},
we sum the expected PG rewards over all
$N_i$ sentences in a fixed reasoning
trajectory $\boldsymbol O_i$:
\begin{equation}
\begin{aligned}
\mathbb E_{\mathbf y^*\sim\eta_H}
\left[
R_i^{\mathrm{PG}}\mid\boldsymbol O_i
\right]
&=
\sum_{j=1}^{N_i}
\mathbb E_{\mathbf y^*\sim\eta_H}
\left[
r_{i,j}^{\mathrm{PG}}\mid\boldsymbol O_i
\right]\\
&=
\frac{1}{L}\sum_{j=1}^{N_i}
\left[
D_{\mathrm{KL}}(\eta_H\Vert p_{j-1}^i)
-
D_{\mathrm{KL}}(\eta_H\Vert p_j^i)
\right]\\
&=
\frac{1}{L}\left[
D_{\mathrm{KL}}(\eta_H\Vert p_0^i)
-
D_{\mathrm{KL}}(\eta_H\Vert p_{N_i}^i)
\right].
\end{aligned}
\label{eq:proof_cumulative_kl}
\end{equation}

Thus, the cumulative expected PG exactly
characterizes the total predictive KL reduction
from the history-only distribution to the
final reasoning-conditioned distribution.
Since $p_0^i$ depends only on $H$,
taking the expectation over complete
reasoning trajectories gives
\begin{equation}
\begin{aligned}
\mathbb E_{\boldsymbol O_i,\mathbf y^*}
[R_i^{\mathrm{PG}}]
=
\frac{1}{L}\left[
D_{\mathrm{KL}}(\eta_H\Vert p_0^i)
-
\mathbb E_{\boldsymbol O_i}
\left[
D_{\mathrm{KL}}(\eta_H\Vert p_{N_i}^i)
\right]
\right].
\end{aligned}
\label{eq:proof_expected_cumulative_kl}
\end{equation}

Finally, applying Lemma A.1 to the final
prediction distribution and using Jensen's
inequality, we obtain
\begin{equation}
\begin{aligned}
\mathbb E_{\boldsymbol O_i}
[\mathcal R_K(p_{N_i}^i)]
&\leq
\mathbb E_{\boldsymbol O_i}
\left[
\sqrt{
2D_{\mathrm{KL}}(\eta_H\Vert p_{N_i}^i)
}
\right]\\
&\leq
\sqrt{
2\mathbb E_{\boldsymbol O_i}
\left[
D_{\mathrm{KL}}(\eta_H\Vert p_{N_i}^i)
\right]
}\\
&=
\sqrt{
2\Big[
D_{\mathrm{KL}}(\eta_H\Vert p_0^i)
-
\mathbb E_{\boldsymbol O_i,\mathbf y^*}
[L R_i^{\mathrm{PG}}]
\Big]
}.
\end{aligned}
\label{eq:proof_cumulative_final}
\end{equation}
This establishes
Eq.~\ref{eq:theory_cumulative_bound}
and completes the proof.
\hfill$\square$

\subsection{Training Computation Budget Analysis}
\label{sec:flops_analysis}
In this section, we theoretically analyze the additional
FLOPs of GainReason compared with vanilla GRPO.
Since GainReason follows the vanilla GRPO training pipeline,
we focus on the additional forward computation required
for sentence-level PG estimation.
\subsubsection{\textbf{FLOPs of GRPO}}
FLOPs (Floating-Point Operations) measure the number
of arithmetic operations required by a computation.
We focus on matrix multiplication, which dominates
the computational cost of Transformers.
For matrices $A\in\mathbb{R}^{m\times k}$ and
$B\in\mathbb{R}^{k\times p}$, their multiplication
requires approximately $2mkp$ FLOPs. 

Let $b$, $s$, $h$, $l$, and $V$ denote the batch size,
sequence length, hidden dimension, number of Transformer
layers, and vocabulary size, respectively.
Following the Transformer FLOPs analysis in~\cite{igpo},
the FLOPs of a single GRPO forward pass are:
\begin{equation}
\begin{aligned}
F_{\mathrm{GRPO}}
&=24lbsh^2+4lbs^2h+2bshV\\
&=\alpha s^2+\beta s,
\end{aligned}
\label{eq:grpo_flops}
\end{equation}
where $\alpha=4lbh$ and
$\beta=24lbh^2+2bhV$.

\subsubsection{\textbf{Additional FLOPs of GainReason}}

Our custom causal attention mechanism  avoids repeated forward
passes over shared reasoning prefixes by extending
the original sequence with additional SID blocks.
Let $t$ denote the text length,
$\ell_{\mathrm{SID}}$ the target SID length,
and $N$ the number of reasoning sentences.
The original GRPO sequence has length
$s=t+\ell_{\mathrm{SID}}$, whereas GainReason
extends it to:
\begin{equation}
s_{\mathrm{PG}}
=t+(N+1)\ell_{\mathrm{SID}}
=s+N\ell_{\mathrm{SID}}.
\label{eq:igrec_seq_length}
\end{equation}

For simplicity, let
$g=N\ell_{\mathrm{SID}}$ denote the number
of additional SID tokens.
The FLOPs of the extended forward pass are:
\begin{equation}
F_{\mathrm{GainReason}}
=\alpha(s+g)^2+\beta(s+g).
\label{eq:igrec_flops}
\end{equation}

Therefore, the additional FLOPs compared with GRPO are:
\begin{equation}
\begin{aligned}
\Delta F
&=F_{\mathrm{GainReason}}-F_{\mathrm{GRPO}}\\
&=2\alpha sg+\alpha g^2+\beta g.
\end{aligned}
\label{eq:igrec_delta_flops}
\end{equation}

The relative FLOPs overhead is then:
\begin{equation}
\begin{aligned}
\delta
&=\frac{\Delta F}{F_{\mathrm{GRPO}}}
=\frac{g(2\alpha s+\alpha g+\beta)}
        {\alpha s^2+\beta s}\\
&=\frac{g}{s}
\left(1+\frac{s+g}{s+\beta/\alpha}\right)
=\frac{g}{s}
\left(1+\frac{s+g}{s+6h+\frac{V}{2l}}\right).
\end{aligned}
\label{eq:igrec_relative_flops}
\end{equation}

Let $N_{\max}>1$ denote the maximum number of
reasoning sentences, and define
$g_{\max}=N_{\max}\ell_{\mathrm{SID}}$.
In LLM-based GR, $g$ is typically much smaller than
$6h+\frac{V}{2l}$. 
Consequently, the relative FLOPs overhead can be bounded as:
\begin{equation}
\begin{aligned}
\delta
< \frac{2g}{s}
\leq \frac{2N_{\max}\ell_{\mathrm{SID}}}{s}.
\end{aligned}
\label{eq:igrec_flops_bound}
\end{equation}

This bound shows that the additional FLOPs introduced
by  PG estimation with custom causal attention are controlled by the
number of reasoning sentences and the target SID length.
Since $\ell_{\mathrm{SID}}$ is typically small,
the computational overhead remains bounded under
the adopted reasoning segmentation strategy.
We further evaluate the practical training overhead
in Section~\ref{sec:compute_analysis}.

\section{Additional Experimental Details}
\subsection{Implementation Details}
\label{sec:implementation}
\textbf{SID Construction.} 
 For Movies and Yelp, we encode item metadata using Sentence-T5 and apply three-level RQ-KMeans~\cite{onerec} to construct SIDs. At each level, K-means clusters the residual embeddings, and the selected centroid is subtracted before quantization at the next level. The three resulting cluster indices form each item's SID. We use 256 centroids per level for Movies and Yelp. For Industrial, we directly reuse the publicly released SIDs from SIDReasoner~\cite{sidreasoner}.

 \textbf{Itemic Alignment.}
We initialize the model from Qwen3-1.7B~\cite{qwen3} and extend its vocabulary with the SID tokens constructed above. Following prior works~\cite{onerec-think,sidreasoner,lcrec}, we perform multi-task itemic alignment using AdamW with a batch size of 128 and a learning rate of $1\times10^{-4}$. To reduce overfitting to specific prompt formulations, we use at least four prompt variants for each task. Section~\ref{sec:prompts} presents one representative template per task.

\textbf{Reasoning Activation.}
After itemic alignment, we activate the model's reasoning format through SFT. For each dataset, we use Qwen3-8B~\cite{qwen3} to generate rationales offline for the training instances. The aligned model is fine-tuned with a batch size of 128 and a learning rate of $2\times10^{-5}$.

\textbf{Reasoning Optimization.}  
After reasoning activation, the model is further optimized on the full training set. Our codes are  based on verl~\cite{verl}. Following prior works~\cite{onerec-think}, we train for 2 epochs, with a training batch size of 480, a micro-batch size of 48, and a learning rate of $5\times10^{-7}$. Rollouts are generated using vLLM with a temperature of 1.0, sampling 8 trajectories per input with a maximum response length of 512 tokens. To ensure a controlled comparison of RL strategies, all explicit reasoning-based baselines and GainReason are initialized from the same checkpoint obtained after itemic alignment and reasoning activation and share these training settings.  For GainReason, we tune the PGRU weighting coefficient
$\lambda$ (Eq.~\ref{eq:IG-Rec_objective}) over
$\{0.01, 0.1, 0.3, 0.5, 0.8\}$ and select 0.5.
The utilization range $W$ (Eq.~\ref{eq:igru_weight}) is tuned over
$\{0.1, 0.2, 0.3, 0.4, 0.5\}$, with 0.2 selected.
We initially treat each sentence as a reasoning step.
For rollouts exceeding 20 segments, we iteratively
merge the shortest segment with an adjacent segment
until at most 20 segments remain. All training experiments are conducted on eight NVIDIA A100 80GB GPUs.  



\subsection{Baseline Methods}
\label{sec:baselines}
\subsubsection{\textbf{Traditional Sequential Recommenders}}
\begin{itemize}
    \item \textbf{GRU4Rec}~\cite{gru4rec} uses gated recurrent units to encode interaction sequences and predict the next item from the evolving hidden state.

    \item \textbf{SASRec}~\cite{sasrec} employs causal self-attention to capture dependencies in interaction histories and predicts the next item with a cross-entropy objective.

    \item \textbf{BERT4Rec}~\cite{bert4rec} uses a bidirectional Transformer to model interaction sequences and is trained to reconstruct masked items with a cross-entropy objective.
\end{itemize}

\subsubsection{\textbf{Generative Recommenders}}
\begin{itemize}
    \item \textbf{TIGER}~\cite{tiger} constructs SIDs using RQ-VAE and trains an encoder--decoder Transformer to autoregressively generate the next item's SID with a cross-entropy objective.

    \item \textbf{RPG}~\cite{rpg} represents items with long, unordered SIDs and predicts their tokens in parallel using a multi-token prediction objective. It employs graph-guided decoding to retrieve valid items efficiently.

    \item \textbf{LLaDA-Rec}~\cite{lladarec} formulates SID generation as discrete diffusion, combining parallel item tokenization with bidirectional masked prediction. It uses adaptive-order decoding with an adapted beam search strategy.

     \item \textbf{LC-Rec}~\cite{lcrec} adapts a pretrained LLM for SID generation through sequential recommendation and auxiliary SID--language alignment tasks. For a controlled comparison, we use the same training inputs for model preparation. 
\end{itemize}

\subsubsection{\textbf{Reasoning-Enhanced Generative Recommenders}}
\begin{itemize}
    \item \textbf{ReaRec}~\cite{rearec} recursively feeds the final hidden state back into the  recommender, progressively refining user representations through multi-step latent reasoning. 

    \item \textbf{CARE}~\cite{care} mitigates bias amplification in SID generation through progressive history encoding and query-anchored latent reasoning. It incorporates increasingly fine-grained historical information and learnable queries at successive token-generation stages.
    

    \item \textbf{OneRec-Think}~\cite{onerec-think} performs itemic alignment to align SIDs with the language representation space, followed by reasoning activation to establish explicit reasoning patterns. It then enhances reasoning through GRPO with a Rollout-Beam reward, which evaluates whether the target SID appears among beam-search candidates conditioned on the generated rationales.

    \item \textbf{SIDReasoner}~\cite{sidreasoner} adopts itemic alignment and reasoning activation stages similar to OneRec-Think. It subsequently employs GRPO with prefix-based SID correctness and format rewards to optimize reasoning based on final prediction outcomes.

    \item \textbf{Evo-Rec}~\cite{evo-rec} adopts an itemic alignment stage similar to OneRec-Think, followed by Best-of-N rejection sampling SFT for reasoning activation. It then applies ranking-aware GRPO, using the NDCG of the target SID under beam search as the trajectory-level reward.

    \item \textbf{SAPO}~\cite{sapo} pairs each reasoning block with a corresponding SID token and assigns block-level advantages based on token-wise SID correctness, with length-normalized updates within each block. We implement SAPO ourselves following the original paper.
\end{itemize}

\section{Additional Training Details}
\label{sec: training_details}
\subsection{Itemic Alignment Task}
We perform SFT with two categories of itemic alignment tasks: 
\begin{itemize}
    \item \textbf{Item prediction.}
    The model predicts the next item from a user's interaction history using three task variants: (1) \emph{SID-based prediction}, which maps historical SIDs to the target SID and serves as the basic recommendation objective; (2) \emph{text-based prediction}, which maps historical item titles to the target title; and (3) \emph{cross-representation prediction}, which maps historical SIDs to the target title or historical titles to the target SID. These tasks capture behavioral patterns, with the cross-representation variant explicitly requiring the model to connect SID representations with item semantics.
    
    \item \textbf{Item-SID translation.}
    The model learns bidirectional mappings between item titles and their corresponding SIDs, generating a title from an SID sequence and vice versa. This task grounds SID tokens in the LLM's linguistic space.
\end{itemize}

The prompts are illustrated in Section~\ref{sec:prompts}.

\subsection{Reasoning Activation}
Despite the substantial recommendation improvements achieved through itemic alignment, we observe that the model still suffers from format drift and struggles to generate structurally consistent and semantically coherent reasoning at inference time, limiting its reasoning effectiveness and interpretability. To address this issue, following prior works~\cite{sidreasoner,onerec-think}, we introduce a reasoning activation stage. Specifically, we construct offline CoT trajectories following the principles of OneReason~\cite{onereason}, where each trajectory summarizes user behavior and preferences, explores plausible interest directions, and weighs behavioral evidence to infer the next purchase (see Section~\ref{sec:reasoning_construction} for details). We then perform SFT on the training instances with the constructed rationales. This stage establishes a structured reason-then-recommend generation pattern, where the model first generates a reasoning trajectory and subsequently predicts the next SID conditioned on the interaction history and the generated reasoning.

\subsection{Prompt Templates}
\label{sec:prompts}
\subsubsection{\textbf{Itemic Alignment}}

\begin{magicprompt}{SID-based Prediction}
<|im_start|>system
You are a helpful assistant.
<|im_end|>
<|im_start|>user
The user has interacted with items {history_sid} in chronological order. Can you predict the next possible item that the user may expect?<|im_end|>
<|im_start|>assistant
<think>
</think>

<|sid_begin|>{target_sid}<|sid_end|><|im_end|>
\end{magicprompt}

\begin{magicprompt}{Text-based Prediction}
<|im_start|>system
You are a helpful assistant.
<|im_end|>
<|im_start|>user
Given the user's historical interactive items arranged in chronological order: {history_titles}, can you recommend a suitable item for the user?<|im_end|>
<|im_start|>assistant
<think>

</think>

{target_title}
<|im_end|>
\end{magicprompt}

\begin{magicprompt}{Fusion Prediction (SID--Title)}
<|im_start|>system
You are a helpful assistant.
<|im_end|>
<|im_start|>user
After interacting with items {history_sid}, what is the most probable item for the user to interact with next? Kindly provide the item's title.<|im_end|>
<|im_start|>assistant
<think>

</think>

{target_title}<|im_end|>
\end{magicprompt}

\begin{magicprompt}{Fusion Prediction (Title--SID)}
<|im_start|>system
You are a helpful assistant.
<|im_end|>
<|im_start|>user
Considering the title sequence of user history interaction items: "{history_titles}". What is the next recommendation for the user?<|im_end|>
<|im_start|>assistant
<think>

</think>

<|sid_begin|>{target_sid}<|sid_end|><|im_end|>
\end{magicprompt}

\begin{magicprompt}{Item--SID Translation (Title--SID)}
<|im_start|>system
You are a helpful assistant.
<|im_end|>
<|im_start|>user
An item is called "{item_title}", can you tell me which item it is?<|im_end|>
<|im_start|>assistant
<think>

</think>

<|sid_begin|>{item_sid}<|sid_end|><|im_end|>
\end{magicprompt}

\begin{magicprompt}{Item--SID Translation (SID--Title)}
<|im_start|>system
You are a helpful assistant.
<|im_end|>
<|im_start|>user
Could you please tell me what item {item_sid} is called?<|im_end|>
<|im_start|>assistant
<think>

</think>

{item_title}<|im_end|>
\end{magicprompt}

\subsubsection{\textbf{Reasoning Training}}

\begin{magicprompt}{}
<|im_start|>system
You are a helpful assistant.
<|im_end|>
<|im_start|>user
The user has sequentially interacted with items {history_sid}. Can you recommend the next item for him? Let's think step by step before making recommendation. When reasoning, output only <Step>...</Step> blocks. After thinking, directly output the item semantic IDs.<|im_end|>
<|im_start|>assistant
<think>
{reasoning_text}
</think>

<|sid_begin|>{target_sid}<|sid_end|><|im_end|>
\end{magicprompt}

\subsubsection{\textbf{Reasoning Construction}}
\label{sec:reasoning_construction}
\begin{magicprompt}{Amazon Datasets}
You are an expert in reasoning about user shopping behavior for e-commerce recommendation. Given a user's chronological purchase history, infer the user's current shopping interests and plausible future needs through concise, evidence-grounded reasoning.

=== USER PURCHASE HISTORY ===

{history_metadata}

=== SUBSEQUENT INTERACTION (ANSWER; MUST NOT APPEAR EXPLICITLY IN THE REASONING) ===

{target_metadata}

The subsequent interaction is provided only during rationale construction. Write the reasoning from the perspective that it has not happened yet and the user's actual next purchase is unknown. Never mention, acknowledge, compare against, or explicitly describe the subsequent interaction. It represents only one possible realized future, not the uniquely correct outcome.

The following three stages are a reasoning reference rather than a mandatory checklist. Use only the stages that are supported by the available evidence, and reduce the number of <Step> blocks when the history is too sparse to justify deeper reasoning. Do not invent conclusions merely to complete all three stages.

**Step 1 — Shopping Behavior Summary:**

Summarize the strongest signals in the purchase history, such as recent focus, repetition, shared specifications, meaningful product relations, or changes in interest. When evidence is sparse, keep the conclusion local rather than inferring stable preferences or recurring needs.

**Step 2 — Plausible Possibility Expansion:**

When supported by the history, explore several plausible next-purchase directions, such as continuation, replenishment, complementarity, substitution, specification change, task continuation, exploration, or interest shift. Reason divergently before committing to one direction, but do not create unnecessary possibilities.

**Step 3 — Evidence-Weighted Next-Purchase Inference:**

When sufficient evidence is available, compare the plausible directions using repetition, recency, specificity, temporal continuity, coherent product relations, and required assumptions. Infer the most plausible forward-looking shopping need or direction while retaining uncertainty when the evidence is weak. Do not predict an exact product.

**Requirements:**

- Write in English only and output only <Step>...</Step> blocks.
- Use one to three <Step> blocks according to the amount of useful evidence.
- Never mention the subsequent interaction, answer, target, or the fact that future information was provided.
- Never reveal its title, brand, exact specifications, or other uniquely identifying details.
- Do not retrospectively reinterpret the history merely to make the actual future interaction appear inevitable.
- Do not invent unsupported user scenarios, motivations, stable preferences, or temporal patterns.
- Keep each Step concise, case-specific, and information-dense.
\end{magicprompt}

\begin{magicprompt}{Yelp Dataset}
You are an expert in reasoning about user preferences for local business recommendation on Yelp. Given a user's chronological interaction history, infer the user's current interests and plausible future interaction directions through concise, evidence-grounded reasoning.

=== USER INTERACTION HISTORY ===

{history_metadata}

=== SUBSEQUENT INTERACTION (ANSWER; MUST NOT APPEAR EXPLICITLY IN THE REASONING) ===

{target_metadata}

The subsequent interaction is provided only during rationale construction. Write the reasoning from the perspective that it has not happened yet and the user's actual next interaction is unknown. Never mention, acknowledge, compare against, or explicitly describe the subsequent interaction. It represents only one possible realized future, not the uniquely correct outcome.

The following three stages are a reasoning reference rather than a mandatory checklist. Use only the stages that are supported by the available evidence, and reduce the number of <Step> blocks when the history is too sparse to justify deeper reasoning. Do not invent conclusions merely to complete all three stages.

**Step 1 — Interaction Preference Summary:**

Summarize the strongest signals in the interaction history, such as recent focus, repeated business categories, cuisine or service preferences, geographic patterns, price or ambience tendencies, meaningful business relations, or changes in interest. A single historical interaction does not establish a stable user preference; in sparse histories, describe only directly supported tendencies and avoid treating incidental business attributes as preferences.

**Step 2 — Plausible Possibility Expansion:**

When supported by the history, explore several plausible next-interaction directions, such as continuing a recent category, revisiting a familiar business type, exploring related cuisines or services, choosing businesses with similar location, price, ambience, or service characteristics, or shifting toward another emerging interest. Every proposed direction must be independently supported by the interaction history rather than introduced merely because it matches the subsequent interaction. Reason divergently before committing to one direction, but do not create unnecessary possibilities.

**Step 3 — Evidence-Weighted Next-Interaction Inference:**

When sufficient evidence is available, compare the plausible directions using repetition, recency, specificity, category consistency, geographic continuity, coherent business attributes, and required assumptions. Infer the most plausible forward-looking preference or business-interaction direction while retaining uncertainty when the evidence is weak. Use this stage only when the history provides enough evidence to meaningfully rank competing directions. Do not predict an exact business.

**Requirements:**

- Write in English only and output only <Step>...</Step> blocks.
- Use one to three <Step> blocks according to the amount and strength of useful evidence.
- With only one historical interaction, normally use one or two Steps unless there is sufficient evidence for a meaningful comparison.
- Never mention the subsequent interaction, answer, target, or the fact that future information was provided.
- Never reveal its business name, exact location, categories, attributes, or other uniquely identifying details.
- Every inferred pattern, preference, and possibility must be supportable from the history alone.
- Do not retrospectively reinterpret the history merely to make the actual future interaction appear likely or inevitable.
- Do not invent unsupported user scenarios, motivations, stable preferences, geographic habits, or temporal patterns.
- Keep each Step concise, case-specific, and information-dense.
\end{magicprompt}

\end{document}